%% file: main.tex
\documentclass{article}

\usepackage{microtype}
\usepackage{graphicx}
\usepackage{booktabs} 

\usepackage{footmisc}
\usepackage{multirow}
\usepackage[caption=false,font=footnotesize]{subfig}
\usepackage{textcomp}
\usepackage{xcolor}
\graphicspath{{./figures/}}
\usepackage{url}
\usepackage{hyperref}

\usepackage{adjustbox}

\usepackage{cite}
\usepackage{amsmath,amssymb,amsfonts}

\usepackage[accepted]{icml2021}

\icmltitlerunning{Generative Causal Explanations for Graph Neural Networks}

\begin{document}

\twocolumn[
\icmltitle{Generative Causal Explanations for Graph Neural Networks}

\icmlsetsymbol{equal}{*}

\begin{icmlauthorlist}
\icmlauthor{Wanyu Lin}{hk,toronto}
\icmlauthor{Hao Lan}{toronto}
\icmlauthor{Baochun Li}{toronto}
\end{icmlauthorlist}

\icmlaffiliation{hk}{Department of Computing, The Hong Kong Polytechnic University, Hong Kong, China}
\icmlaffiliation{toronto}{Department of Electrical \& Computer Engineering, University of Toronto, Toronto, Canada}

\icmlcorrespondingauthor{Wanyu Lin}{wanylin@comp.polyu.edu.hk}

\icmlkeywords{Model Explanation, Graph Neural Networks, Granger Causality}

\vskip 0.3in
]
\printAffiliationsAndNotice{}

\begin{abstract}

This paper presents {\em Gem}, a model-agnostic approach for providing interpretable explanations for any GNNs on various graph learning tasks. Specifically, we formulate the problem of providing explanations for the decisions of GNNs as a causal learning task. Then we train a causal explanation model equipped with a loss function based on Granger causality. Different from existing explainers for GNNs, {\em Gem} explains GNNs on graph-structured data from a causal perspective. It has better generalization ability as it has no requirements on the internal structure of the GNNs or prior knowledge on the graph learning tasks. In addition, {\em Gem}, once trained, can be used to explain the target GNN very quickly. Our theoretical analysis shows that several recent explainers fall into a unified framework of {\em additive feature attribution methods}. Experimental results on synthetic and real-world datasets show that {\em Gem} achieves a relative increase of the explanation accuracy by up to $30\%$ and speeds up the explanation process by up to $110\times$ as compared to its state-of-the-art alternatives. 

\end{abstract}

\input{intro}
\input{problem}

\input{solution}

\input{experiments}

\input{related}

\input{conclusion}

\input{ack}

\bibliography{main}
\bibliographystyle{icml2021}


\end{document}


\twocolumn[
\icmltitle{Generative Causal Explanations for Graph Neural Networks}

\icmlkeywords{Model Explanation, Graph Neural Networks, Granger Causality}

\vskip 0.3in
]

\appendix
\blfootnote{The source code can be found in https://github.com/wanyu-lin/ICML2021-Gem.}

\input{appendix}

\bibliography{main}
\bibliographystyle{icml2021}

%% file: intro.tex
\section{Introduction}
\label{sec:intro}

Many problems in scientific domains, ranging from social networks~\cite{wanyu-infocom20} to biology~\cite{zitnik2018modeling} and chemistry~\cite{zitnik2017predicting}, can be naturally modeled as problems of property learning on graphs. For example, in biology, identifying the functionality of proteins is critical to find the proteins associated with a disease, where proteins are represented by local protein-protein interaction (PPI) graphs. Supervised learning of graphs, especially with graph neural networks (GNNs), has had a significant impact on these domains, mainly owing to their efficiency and capability of inductive learning~\cite{hamilton2017inductive}.

Despite their practical success, most GNNs are deployed as black boxes, lacking explicit declarative knowledge representations. Therefore, they have difficulty in generating the required underlying explanatory structure. The deficiency of explanations for the decisions of GNNs significantly hinders the applicability of these models in decision-critical settings, where both predictive performance and interpretability are of paramount importance. For example, medical decisions are increasingly being assisted by complex predictions that should lend themselves to be verified by human experts easily. Model explanations allow us to argue for model decisions and exhibit the situation when algorithmic decisions might be biased or discriminating. In addition, precise explanations may facilitate model debugging and error analysis, which may help decide which model would better describe the data's underlying semantics.

While explaining graph neural networks on graphs is still a nascent research topic, a few recent works have emerged~\cite{ying2019gnnexplainer, yuan2020xgnn,luo2020parameterized,vu2020pgm}, each with its own perspective on this topic. In particular, XGNN~\cite{yuan2020xgnn} was proposed to investigate graph patterns that lead to a specific class, while GNNExplainer~\cite{ying2019gnnexplainer} provided the local explanation for a single instance (a node/link/graph), by determining a compact subgraph leading to its prediction. PGM-Explainer~\cite{vu2020pgm} explored the dependencies of explained features in the form of conditional probability, which is naturally designed for explaining a single instance.  

However, verifying if a target GNN works as expected often requires a considerable amount of explanations for providing a global view of explanations. For this end, PGExplainer~\cite{luo2020parameterized} learns a multilayer perceptron (MLP) to explain multiple instances collectively. However, PGExplainer heavily relies on node embeddings from the target GNN, which may not be obtained without knowing its internal model structure and parameters. Besides, PGExplainer can not explain any graph tasks without explicit motifs. Taking MUTAG as an example, PGExplainer assumes that $\textrm{NH}_2$ and $\textrm{NO}_2$ are the motifs for the mutagen graphs and filters out the instances without these two motifs. We verified that the assumption might not be reasonable by looking at the dataset statistics (provided in Appendix B). More specifically, PGExplainer fails to explain the instances without explicit motifs under their assumptions. Motivated by this observation, we aim to provide fast and accurate explanations for any GNN-based models without the limitations above.


In this work, we propose a new methodology, called {\em Gem}, to provide interpretable explanations for any GNNs on the graph using causal explanation models. To the best of our knowledge, while the notion of causality has been used for interpretable machine learning on images or texts, this is the first effort from a causal perspective to explain graph neural networks. Specifically, our causal objective is built upon the notion of Granger causality, which comes from the pioneering work of Wiener and Granger~\cite{granger1969investigating,bressler2011wiener,wiener1956theory}. Granger causality declares a causal relationship $x_i \rightarrow Y$ between variables $x_i$ and $Y$ if we are better able to predict $Y$ using all available information than if the information apart from $x_i$ had been used. In the graph domain, if the absence of an edge/node $x_i$ decreases the ability to predict $Y$, then there is a causal relationship between this edge/node and its corresponding prediction. Based on the insights from neuroscience~\cite{biswal1997simultaneous}, we extend the notion of Granger causality to characterize the explanation of an instance by its local subgraphs.

We note that the concept of Granger causality is probabilistic, and the graph data is inherently interdependent, i.e., edges or nodes are correlated variables. Directly applying Granger causality may lead to incorrectly detected causal relations. In addition, we envision that the resulting explanations should be human-intelligible and valid. For example, in some applications such as chemistry, an explanation for the mutagen graph is a functional group and should be connected. Accordingly, we propose an approximate computation strategy that makes our method viable for graph data with interdependency, under reasonable assumptions on the causal objective. 

In particular, we incorporate various graph rules, such as the connectivity check, to encourage the obtained explanations to be valid and human-intelligible. Then we train causal explanation models that learn to distill compact subgraphs, causing the outputs of the target GNNs. This approach is flexible and general since it has no requirements on the target model to be explained (commonly referred to as ``model-agnostic''), or no assumptions on the learning tasks (explicit motifs for identifying a particular class), and can provide local and global views of the explanations. In particular, it does not require retraining or adapting to the original model. In other words, once trained, {\em Gem} can be used to explain the target GNN models with little time. 

Highlights of our original contributions are as follows. We propose a new methodology to explain graph neural networks on the graph from the causal perspective; to the best of our knowledge, such an approach has not been used for interpreting GNNs on the graph so far. We introduce causal objectives for better estimates of the causal effect in our methodology and provide an approximate computation strategy to deal with graph data with interdependency. Various graph rules are incorporated to ensure that the obtained explanations are valid. We then use a causal objective to train a graph generative model as the explainer, which can automatically explain the target GNNs with little time. We theoretically analyze that several recent methods, including {\em Gem}, all fall into the framework of additive feature attribution methods, which essentially solve the same optimization problem with different approximation methods (provided in Appendix A). We empirically demonstrate that {\em Gem} is significantly faster and more accurate than alternative methods.

%% file: problem.tex
\section{Problem Setup}
\label{sec:prelim}


A set of graphs can be represented as $\mathcal{G}=\left\{G_i\right\}^N_{i=1}$, where $|\mathcal{G}|=N$. Each graph is denoted as $G_i=\left(V_i,\,E_i\right)$, where $E_i$ denotes the edge set and $V_i=\left\{v^i_1,\,v^i_2,\,\cdots,\,v^i_{|V_i|}\right\}$ is the node set of graph $i$. In many applications, nodes are associated with $d$-dimensional node features $X_i=\left\{x^i_1,\,x^i_2,\,\cdots,\,x^i_{|V_i|}\right\}$, $x^i_j\in\,\mathbb{R}^d$. Without loss of generality, we consider the problem of explaining a graph neural network-based classification task. This task can be node classification or graph classification. For graph classification, we associate each graph $G_i$ with a label, $y_i$, where $y_i\in\,\mathcal{Y}=\left\{c_1,\,c_2,\,\cdots,\,c_l\right\}$, and $l$ is the number of categories. The dataset $\mathcal{D}=\left\{\left(G_i,\,y_i\right)\right\}^N_{i=1}$ is represented by pairs of graph instances and graph labels. Examples of such task include classifying the drug molecule graphs according to their functionality. 

In the node classification setting, each node $v_j\in\,V$ of a graph $G$ is associated with a corresponding node label $y_j\in\,\mathcal{Y}$. Examples of this kind include classifying papers in a citation network, or entities in a social network such as Reddit. The dataset $\mathcal{D}=\left\{\left(v_j, y_j\right)\right\}^{|V|}_{j=1}$ is represented by pairs of nodes and node labels. In general, we use $I_i$ to represent an instance, which is equivalent to $v_i$ for node classification or $G_i$ for graph classification.


{\bf GNN family models.} Graph neural networks (GNNs) are a family of graph message passing architectures that incorporate graph structure and node features to learn a dense representation of a node or the entire graph. In essence, GNNs follow a neighborhood aggregation strategy, where the node representations are updated via iteratively aggregating the representations from its neighbors in the graph. Graph convolutional networks (GCNs) use mean pooling~\cite{kipf2016semi} for aggregation, while GraphSage aggregates the node features via mean/max/LSTM pooling~\cite{hamilton2017inductive}. Taking GCNs as an example, the basic operator for the neighborhood information aggregation is the element-wise $\mathbf{mean}$. After $L$ iterations of aggregation, a node’s representation can capture the structural information within its $L$-hop graph neighborhood.



Formally, a graph neural network (GNN) can be written as a function $f\left(\cdot\right):\,\mathcal{G}\,\rightarrow\,\mathcal{Y}$ or $f\left(\cdot\right):\,V\,\rightarrow\,\mathcal{Y}$. The former is a graph-level classifier, and the latter is a node-level classifier. Typically, a GNN $f\left(\cdot\right)$ is trained with an objective function $\mathcal{L}:\,y\times\,\tilde{y}\,\rightarrow\,s$ that computes a scalar loss $s\in\,\mathbb{R}$ after comparing the model's predictive output $\tilde{y}$ to a ground-truth output $y$. The categorical cross-entropy for classification models is commonly used for such objectives.

{\bf Objective.} We are given a pre-trained classification model, represented by $f\left(\cdot\right)$, and our ultimate goal is to obtain an explanation model, denoted as $f\left(\cdot\right)_{\scriptscriptstyle{\mathbf{exp}}}$, that can provide fast and accurate explanations for the pre-trained model, which can also be called a target GNN. Intrinsically, an explanation is a subgraph that is the most relevant for a prediction --- the outcome of the target GNN, denotes as $\tilde{y}$. Consistent with previous studies in the literature~\cite{yuan2020xgnn}, we focus on explanations on graph structures. In particular, we specifically do not require access to, or knowledge of, the process by which the classification model produces its output, nor do we require the classification model to be differentiable or any specific form. We allow the explainers to retrieve different predictions by performing queries on $f\left(\cdot\right)$.

%% file: solution.tex
\section{Methodology}
\label{sec:method}

\begin{figure}[t]
  \centering
  \includegraphics[scale = 0.31]{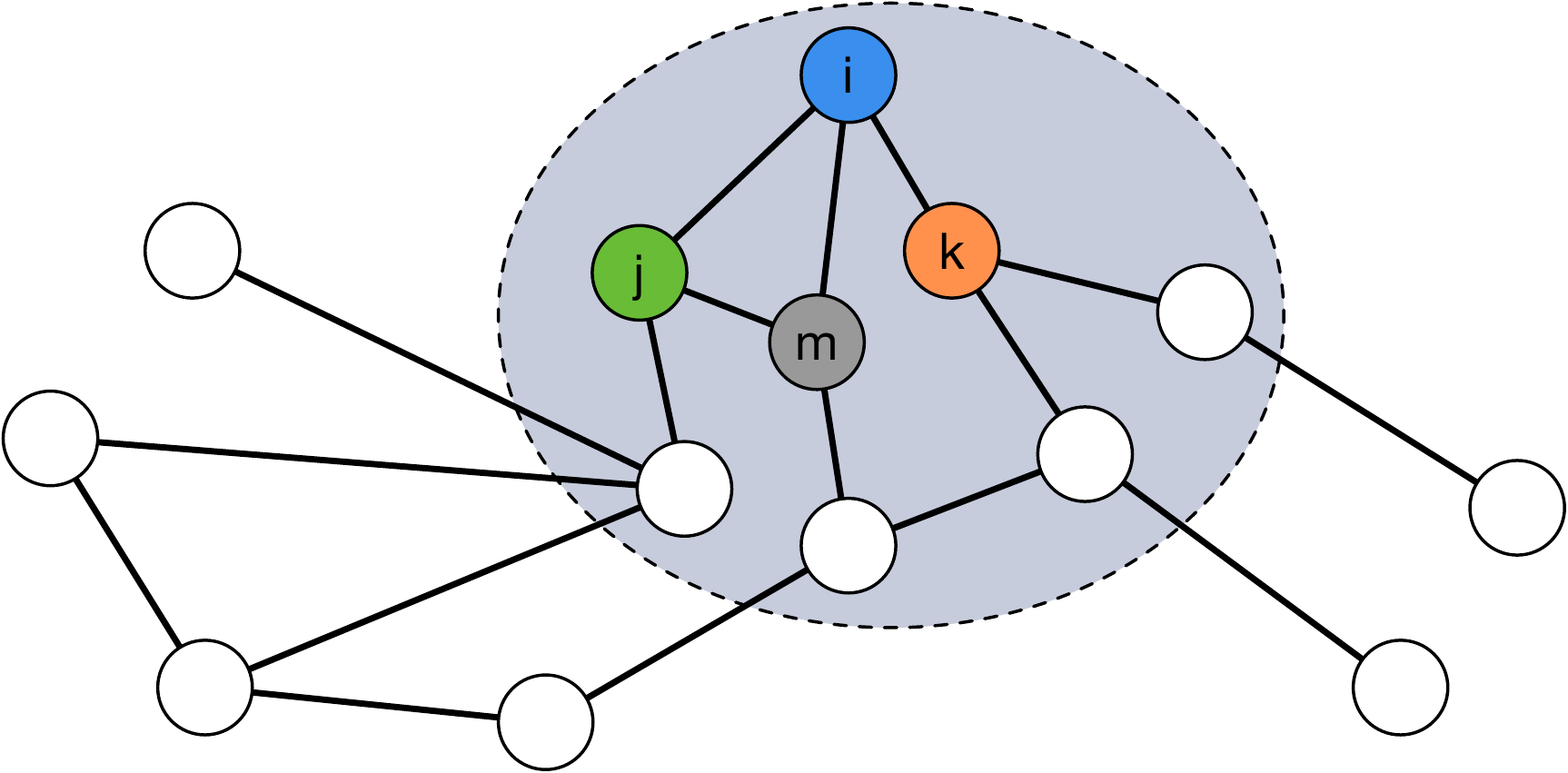}
  \caption{An illustration of the computation graph (best viewed in color). Node $i$ is the target node to be explained.}
  \vspace{-16pt}
  \label{fig:example_graph}
\end{figure}

\begin{figure*}[t]
  \centering
  \includegraphics[scale = 0.2]{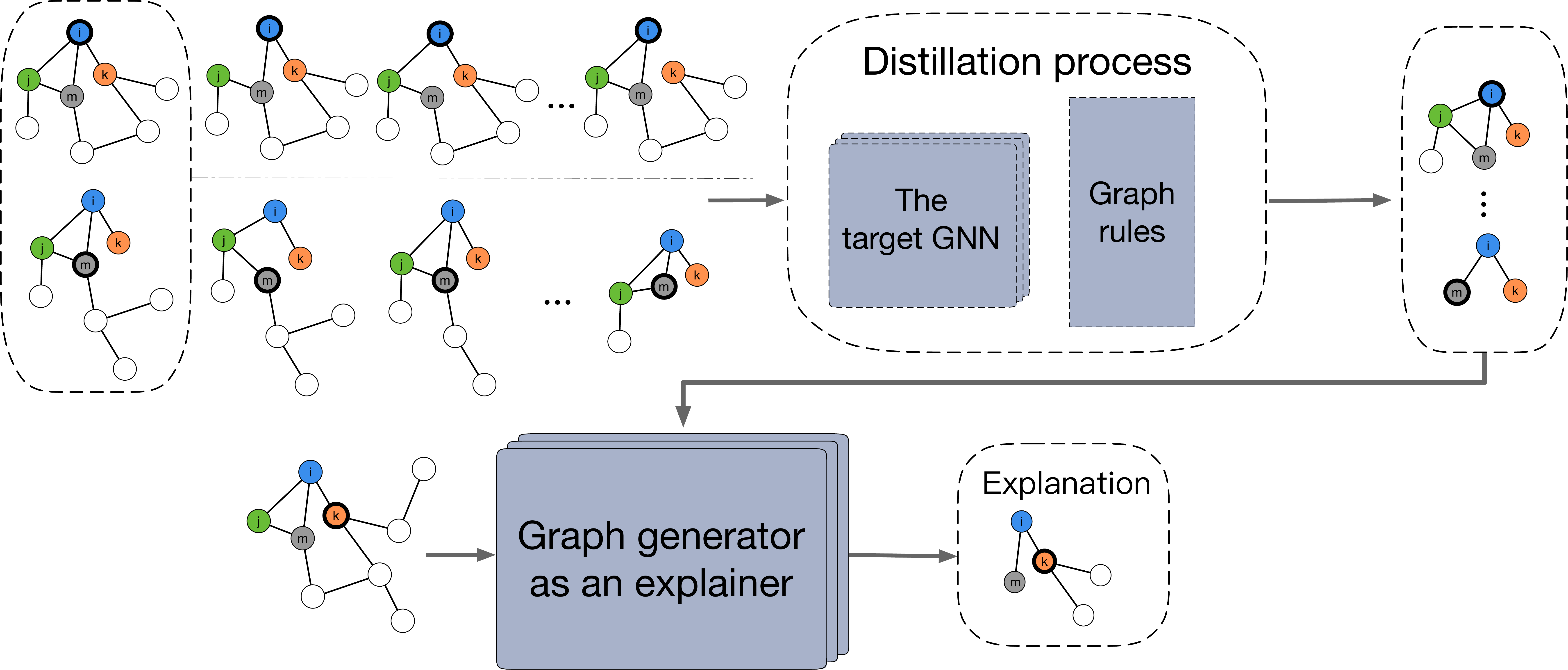}
  \caption{Illustration of {\em Gem}. 1) The distillation process of generating the ground-truth explanations based on the first principles of Granger causality; 2) Training the explainer that can be used to generate explanations for the target GNNs.}
  \vspace{-15pt}
  \label{fig:arc}
\end{figure*}

In essence, the core of the GNNs is a neighborhood-based aggregation process, where a prediction of an instance is fully determined by its computation graph. Let us use $G_i^c=\left(V_i^c,\,A_i^c,\,X_i^c\right)$ to represent the computation graph of an instance $i$, where $V_i^c$ is the node set, $A_i^c\in\,\{0,\,1\}$ indicates the adjacency matrix, and $X_i^c$ is the feature matrix of the computation graph. Typically, a GNN learns a conditional distribution denoted as $\mathcal{P}\left(Y|G_i^c\right)$, where $Y$ is a random variable representing the class labels. For clarity, let us see an example graph, shown in Figure~\ref{fig:example_graph}, which will also be used throughout this paper. In this example, a target GNN is trained for node classification, and the node $i$ is the target node to be explained. Oftentimes, the computation graph of node $i$ is a $L$-hop subgraph; an exmaple of $L=2$ is highlighted in Figure~\ref{fig:example_graph}. 

Therefore, the setting we focus on can be reformulated as the following: we are given a GNN-based classification model that processes the computation graph of an instance (a node or a graph), denoted as $G^c$, and generates the corresponding outputs $p\left(Y|G^c\right)$ for predicting $\tilde{y}$. Unlike the node classification task, when the target GNN is trained for graph classification, the computation graph of an instance will be the entire graph. Accordingly, this work seeks to generate an explanation, a subgraph of $G^c$ that is most relevant for predicting $\tilde{y}$, efficiently and automatically. We use $G^s$ to denote the generated explanation. Our setting is general and works for any graph learning tasks, including node classification and graph classification. Our ultimate goal is to encourage a compact subgraph of the computation graph to have a large causal influence on the outcome of the target GNN. 
 
{\em Differences from PGExplainer.} PGExplainer is the most closely related work to our study, as both PGExplainer and {\em Gem} adopt parameterized networks to provide local and global views for model explanations. However, PGExplainer relies on node embeddings from the target GNN to learn a multilayer perceptron, which may not be obtained without knowing its internal model structure. In contrast, to explain an instance (a node or a graph), {\em Gem} simply inputs the original computation graph into the explainer and outputs a compact explanation graph. In other words, {\em Gem} does not require any prior knowledge of the internal model structure (the target GNN) and parameters, or any prior knowledge of the motifs associated with the graph learning tasks. Therefore, it exhibits better generalization abilities. In what follows, we will present {\em Gem}, our model-agnostic approach for providing interpretable explanations for any GNNs on a variety of graph learning tasks. The design of {\em Gem} is based upon principles of causality, in particular Granger causality~\cite{granger1969investigating}. 


{\bf Granger causality}~\cite{granger1969investigating,granger1980testing}. In general, Granger causality describes the relationships between two (or more) variables when one is causing the other. Specifically, if we are better able to predict variable $\tilde{y}$ using all available information $U$ than if the information apart from variable $x_i$ had been used, we say that $x_i$ Granger-causes $\tilde{y}$~\cite{granger1980testing}, denoted by $x_i\rightarrow \tilde{y}$\footnote{We are aware of the drawbacks of reusing notations. $x_i$ and $\tilde{y}$ in this definition represent any random variables for simplicity.}. 

The crux of our approach is to train an explanation model, or an {\em explainer}, to explain the target graph neural network. Specifically, {\em Gem} is trained with the guidance built on the first principles of Granger causality. Here we extend Granger causality to the case where a compact subgraph $G^s$ of the computation graph is considered the main cause of the corresponding prediction $\tilde{y}$. This is inspired by a long-held belief in neuroscience that the structural connectivity local to a certain area somehow dictates the function of that piece~\cite{biswal1997simultaneous}. Due to the inherent property of GNNs, the computation graph contains all the relevant information that causes the prediction of the target GNN. Under the assumption that Granger causality was built upon, we can squeeze the cause of the prediction, $G^s$, from the computation graph. Therefore, we can use the given definition to quantify the degree to which part of the computation graph causes the prediction of the target GNN. In principle, the notion of Granger causality would hold if $p\left(\tilde{y}\,|\,G^c\right)$ = $p\left(\tilde{y}\,|\,G^s\right)$ holds. 

\subsection{Causal Objective} 
\label{subsec:causal}
Given a pre-trained/target GNN and an instance $G^c$, we use $\delta_{G^c}$ to denote the model error of the target GNN when considering the computation graph, while $\delta_{\scriptscriptstyle{G^c\setminus\left\{e_j\right\}}}$ represents the model error excluding the information from the edge $e_j$, where $e_j\in\,G^c$. With these two definitions and following the notion of Granger causality, we can quantify the causal contribution of an edge $e_j$ to the output of the target GNN. More specifically, the causal contribution of the edge $e_j$ is defined as the decrease in model error, formulated as Eq.~(\ref{eq:loss_diff}):
\begin{equation}\label{eq:loss_diff}
\Delta{\scriptscriptstyle{\delta,\,e_j}}=\delta_{\scriptscriptstyle{G^c\setminus\left\{e_j\right\}}}-\delta_{\scriptscriptstyle{G^c}}
\vspace{-7pt}
\end{equation}

To calculate $\delta_{\scriptscriptstyle{G^c}}$ and $\delta_{\scriptscriptstyle{G^c\setminus\left\{e_j\right\}}}$, we first compute the outputs corresponding to the computation graph $G^c$ and the one excluding edge $e_j$, $G^c\setminus\left\{e_j\right\}$, based on the pre-trained GNN. For simplicity, the pre-trained GNN is denoted as $f\left(\cdot\right)$. Then, the associated outputs can be formulated as Eq.~(\ref{eq:edge}) and Eq.~(\ref{eq:noedge}) respectively:
\begin{equation}\label{eq:edge}
\tilde{y}_{\scriptscriptstyle{G^c}}=f\left(G^c\right)
\vspace{-12pt}
\end{equation}
\begin{equation}\label{eq:noedge}
\tilde{y}_{\scriptscriptstyle{G^c\setminus\left\{e_j\right\}}}=f\left(G^c\setminus\left\{e_j\right\}\right)
\vspace{-7pt}
\end{equation}

Then we compare the outputs of the target GNN, e.g., $\tilde{y}_{\scriptscriptstyle{G^c}}$ and $\tilde{y}_{\scriptscriptstyle{G^c\setminus\left\{e_j\right\}}}$, with the ground-truth label $y$, respectively. In particular, we use the loss function of the pre-trained GNN as the metric to measure the model error, denoted as $\mathcal{L}$. The mathematical formulations are shown as Eq.~(\ref{eq:edge_loss}) and Eq.~(\ref{eq:noedge_loss}):
\begin{equation}\label{eq:edge_loss}
\delta_{\scriptscriptstyle{G^c}} =\mathcal{L}\left(y,\,\tilde{y}_{\scriptscriptstyle{G^c}}\right)
\vspace{-12pt}
\end{equation}
\begin{equation}\label{eq:noedge_loss}
\delta_{\scriptscriptstyle{G^c\setminus\left\{e_j\right\}}}=\mathcal{L}\left(y,\,\tilde{y}_{\scriptscriptstyle{G^c\setminus\left\{e_j\right\}}}\right)
\vspace{-7pt}
\end{equation}

Now, the causal contribution of an edge $e_j$ can be measured by the loss difference associated with the computation graph and the one deleting edge $e_j$.

Recall that we are seeking a ``guidance'' that can be used to train our explainer, encouraging its outcome to be effective explanations. Intrinsically, $\Delta{\scriptscriptstyle{\delta,\,e_j}}$ can be viewed as capturing the individual causal effect (ICE)~\cite{goldstein2015peeking} of the input $G^c$ with values $e_j$ on the output $\tilde{y}$. Therefore, it is straightforward to obtain the most relevant subgraph for predicting $\tilde{y}$ based on $\Delta{\scriptscriptstyle{\delta,\,e_j}}$, which we call the ground-truth distillation process. 

Ideally, given the edges' causal contributions in a computation graph, we can sort the edges accordingly and distill the top-$K$ most relevant edges as a prediction explanation. However, due to the special representations of the graph data, the casual contributions from the edges are not independent, e.g., a $1$-hop neighbor of a node can also be a $2$-hop neighbor of the same node due to cycles. To this end, we further incorporate various graph rules to encourage the distillation process to be more effective. We believe that data characteristics are the most crucial factor in deciding which graph rules to use. It is necessary to understand the principle of the learning task, and the limitation of a human-intelligible explanation might be to prevent spurious explanations. In the application of bioinformatics, such as the MUTAG dataset, the explanation is a functional group, and therefore, the distilled top-$K$ edges should be connected. Nevertheless, in graph representation-based Digital Pathology, such as the cell-graphs towards cancer subtyping, the explanation often contains subsets of cells and cellular interactions~\cite{jaume2020towards}. In this particular scenario, the connectivity constraint is unnecessary. The detailed distillation process is presented in Appendix C. 

Using the distilled ground truth, denoted as $\tilde{G^s}=\left(\tilde{V^s},\,\tilde{A^s},\,\tilde{X^s}\right)$, we can train supervised learning models to learn to explain any other GNN models based solely on its outputs, and without the need to retrain the model to be explained. The workflow of {\em Gem} is illustrated in Figure~\ref{fig:arc}. Note that generating an explanation based on the explainer is not necessary in situations where ground-truth labels of the instances are readily available. In those cases, pre-calculated $\Delta{\scriptscriptstyle{\delta,\,e_j}}$ and our distillation process can directly be used to explain the pre-trained GNN.


\subsection{Graph Generative Model as an Explainer}

In principle, any graph generative models that can be trained to output graph-structured data can be used as a causal explanation model. Due to their simplicity and expressiveness, we focus on auto-encoder architectures utilizing graph convolutions~\cite{kipf2016variational,li2018multi,li2018learning}. In particular, we use a model consisting of a graph convolutional network encoder and an inner product decoder~\cite{kipf2016variational}. We leave the exploration of other generative models for future work.

More concretely, in our explainer, we first apply several graph convolutional layers to aggregate neighborhood information and learn node features. Then we use the inner product decoder to generate an adjacency matrix $\hat{A^c}$ as an explanation mask. With this mask, we are able to construct a corresponding explanation, a compact subgraph that only contains the most relevant portion of a computation graph to its prediction. In particular, each value in $\hat{A^c}$ denotes the contribution of a specific edge to the prediction of $G^c$, if that edge exists in the target instance. Formally, the reconstruction process can be formulated as:
\begin{equation}\label{eq:gcn}
Z=\mathbf{GCNs}\left(A^c, X^c\right)
\vspace{-12pt}
\end{equation}
\begin{equation}\label{eq:reconstruction}
\hat{A^c}=\sigma\left(ZZ^{T}\right)
\vspace{-7pt}
\end{equation}
where $A^c$ is the adjacency matrix of the computation graph for the target instance, $X^c$ denotes the node features, and $Z$ is the learned node features. 

{\bf Explanation for node classification.} The output of an explanation for a target node is a compact subgraph of the computation graph that is most influential for the prediction label. To answer the question of ``How did a GNN predict that a given node has label $\tilde{y}$ ?'', the explainer should capture which node to explain. Specifically, we use a node labeling technique to mark nodes' different roles in a computation graph. The generated node labels are then transformed into vectors (e.g., one-hot encodings) and treated as the nodes' feature vector --- $X^c$.  Note that the labels here represent the structural information of nodes within a computation graph, which are different from the classification/prediction labels. The intuition underlying this node labeling technique is that nodes with different relative positions to the target node may have different structural importance to its prediction label $\tilde{y}$. By incorporating the relative role features, {\em Gem} would capture which node to explain in a computation graph.

\begin{figure}[t]
  \centering
  \includegraphics[scale = 0.31]{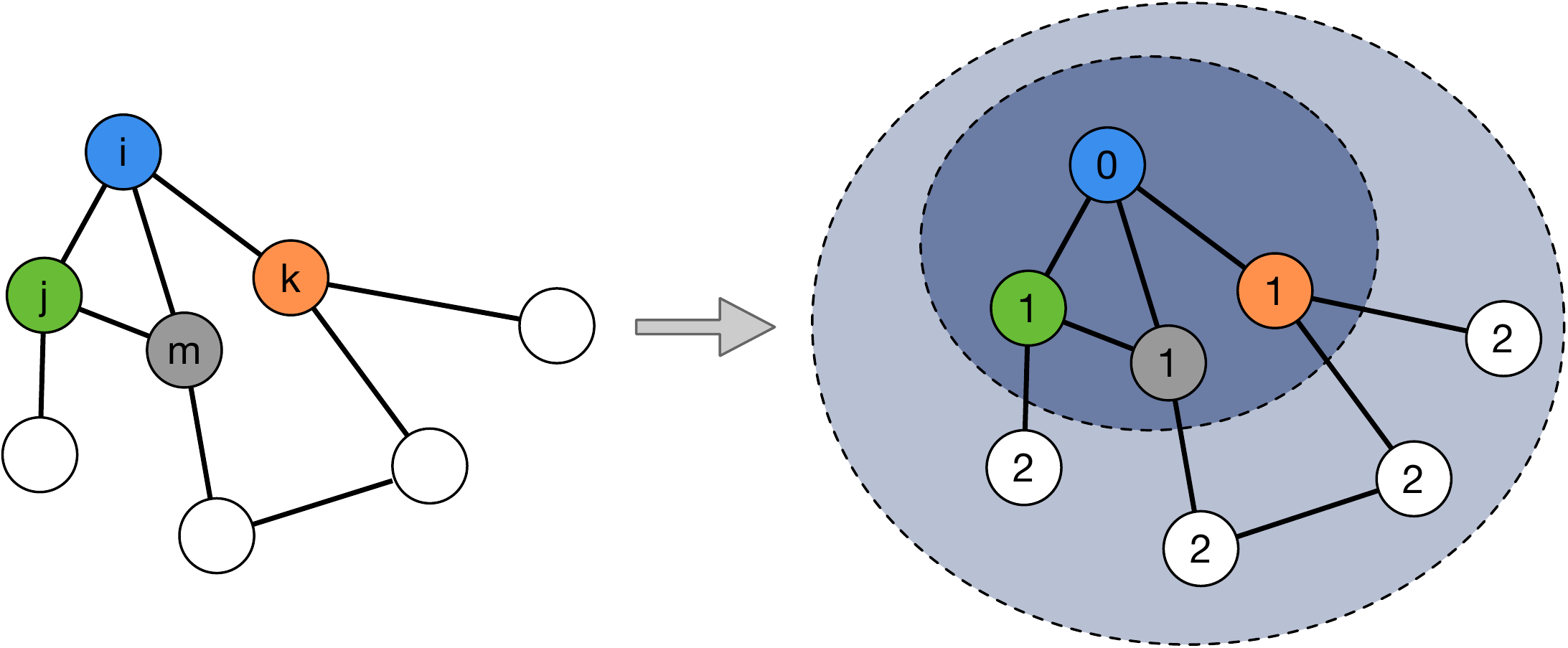}
  \caption{An illustration of the node labeling for node explanation generation (best view in color). The node $i$ is the target node to be explained and the graph in the left-hand side is $i$'s computation graph. $i$ is labeled as $0$, while $j$ and $k$ are labeled as $1$ as they are one-hop away from $i$.}
  \vspace{-18pt}
  \label{fig:exp_node_label}
\end{figure}

Specifically, our node labeling method is derived based on two principles: 1) The target node, denoted as $i$, always has the distinctive label ``$0$.'' 2) A node relative position within a computation graph can be described by its radius regarding the center/target node, i.e., $d(i, j)$ and $d(i, k)$. For two nodes, if $d(i, j) = d(i, k)$, then they have the same label. To further elucidate the used node labeling method, let us see an example shown in Figure~\ref{fig:exp_node_label}. Node $i$ is the target node, while both $j$ and $k$ are the one-hop neighbors of $i$. With our node labeling mechanism, node $i$ is labeled as $0$, while $j$ and $k$ are labeled as $1$ as they are one-hop away from $i$.

{\bf Training the explainer.} We envision that the reconstructed matrix $\hat{A^c}$ is the weighted adjacency matrix that reflects the contributions of the edges to its prediction. Now we can apply the ``guidance'' distilled based on the notion of Granger causality, described in Sec.~\ref{subsec:causal}, to supervise the learning process. In particular, we use the root mean square error between the reconstructed weighted matrices and the true causal contributions distilled based on our proposed distillation process in Sec.~\ref{subsec:causal}.

One highlight of our explainer is its flexibility to choose the predictive model, which is commonly referred to as ``model-agnosticism''~\cite{ribeiro2016model}. Guided by the first principles of Granger causality, our explainer enables graph generative models to learn to generate compact subgraphs for model explanations. We do not need to retrain or adapt the predictive model to explain its decisions. Once trained, it can be used to construct explanations using the generative mapping for the target GNN with little time. 

{\bf Computational complexity analysis.} One may concern that it would be time-consuming to run through the training instances for obtaining the training ``guidance.'' We argue that our method amortizes the estimation cost by training a graph generator to generate explanations for any given instances. In particular, {\em Gem} adopts a parameterized graph auto-encoder with GCN layers to generate explanations, which, once trained, can be utilized in the inductive setting to explain new instances. Specifically, the model parameter complexity of {\em Gem} is independent of the input graph size as it naturally inherits the advantages of GCNs (empirically verified in Appendix B). With the inductive property, the inference time complexity of {\em Gem} is $\mathcal{O}\left(|E|\right)$, where $|E|$ is the number of edges of the instance to be explained. Sec.~\ref{sec:exp} empirically verified the computation efficiency of {\em Gem}. In a nutshell, our solution transforms the task of producing explanations for a given GNN into a supervised learning task, trained based on the first principles of Granger causality. Then we can address the explanation task with existing supervised graph generative models.

{\bf Extensions to other learning tasks on graphs.} Beyond node classification, our explainer can also provide explanations for link prediction and graph classification tasks without modifying its optimization algorithm. The key difference is the node labeling technique for marking the nodes' roles in the computation graph. For example, to generate an explanation for the link prediction task, the explainer model should be able to identify which link to explain. An alternative approach is double-radius node labeling, marking the target link (connecting two target nodes) within the computation graph, proposed by Zhang {\em et al.}~\cite{zhang2018link}. More concretely, node i's position is determined by its radius with respect to the two target nodes $(x,\,y)$, i.e., $\left(d(i,\,x),\,d(i,\,y)\right)$. 

Note that, due to properties of the graph structure invariant, there is no need to mark a particular node/link for graph classification tasks. Instead, we use a graph labeling method, called the Weisfeiler-Lehman (WL) algorithm~\cite{weisfeiler1968reduction}, to capture the structural roles of nodes within a computation graph, which has been widely used in graph isomorphism checking. For more details about the WL algorithm, We refer curious readers to~\cite{weisfeiler1968reduction}. In Sec.~\ref{sec:exp}, we will empirically show that with the ``guidance'' based on Granger causality, complemented by node features from graph/node labeling, {\em Gem} can provide fast and accurate explanations for any graph learning tasks.


%% file: experiments.tex
\section{Experimental Studies}
\label{sec:exp}
\def\arraystretch{0.8}

\subsection{Datasets and Experimental Settings}
{\bf Node classification with synthetic datasets.} In the node classification setting, we built two synthetic datasets where ground truth explanations are available. In particular, we followed the data processing as reported in GNNExplainer~\cite{ying2019gnnexplainer}. The first dataset is BA-shapes, where nodes are labeled based on their structural roles within a house-structured motif, including ``top-node,'' ``middle-node,'' ``bottom-node,'' and ``none-node'' (the ones not belonging to a house). The second dataset is Tree-cycles, in which nodes are labeled to indicate whether they belong to a cycle, including ``cycle-node'' and ``none-node'' (more details of the datasets are provided in Appendix B).

{\bf Graph classification with real-world datasets.} For graph classification, we use two benchmark datasets from bioinformatics --- Mutag~\cite{debnath1991structure} and NCI1~\cite{wale2008comparison}. Mutag contains 4337 molecule graphs, where nodes represent atoms, and edges denote chemical bonds. The graph classes, including the non-mutagenic or the mutagenic class, indicate their mutagenic effects on the Gram-negative bacterium Salmonella typhimurium. NCI1 consists of 4110 instances, each of which is a chemical compound screened for activity against non-small cell lung cancer or ovarian cancer cell lines.

\textit{Baselines.} We consider the state-of-the-art baselines that belong to the unified framework of additive feature attribution methods (The proof is provided in Appendix A)~\cite{lundberg2017unified}: GNNExplainer~\cite{ying2019gnnexplainer} and PGExplainer~\cite{luo2020parameterized}\footnote{We use the source code released by the authors.}. GNNExplainer explains for a given instance at a time, while PGExplainer explains multiple instances collectively. Unless otherwise stated, all the hyperparameters of the baselines are the same in the source code. We do not include gradient-based method~\cite{ying2019gnnexplainer}, graph attention method~\cite{velivckovic2018graph}, and Gradient~\cite{pope2019explainability}, since previous explainers ~\cite{ying2019gnnexplainer,luo2020parameterized} have shown their superiority over these methods.


{\bf Parameter settings of {\em Gem}\footnote{The source code can be found in https://github.com/wanyu-lin/ICML2021-Gem.}.} For all datasets on different tasks, associate explainers share the same structure~\cite{kipf2016variational}. Specifically, we first apply an inference model parameterized by a three-layer GCN with output dimensions $32$, $32$, and $16$. Then the generative model is given by an inner product decoder between latent variables. The explainer models are trained with a learning rate of $0.01$. We use hyperparameter $K$ to control the size of the explanation subgraph and compare the performance of {\em Gem} with GNNExplainer and PGExplainer. The target GNN model accuracy on four datasets, more implementation details, and experimental results are presented in the Appendix B.

{\bf Evaluation metrics.} An essential criterion for explanations is that they must be human interpretable, which implies that the generated explanations should be easy to understand. Taking BA-shapes as an example, the node label is determined by its position within a house-structured motif. The explanations for this dataset should be able to highlight the house structure. For interpretability, we use the visualized explanations of different methods to analyze their performance qualitatively. 

In addition, explanations seek to answer the question: when a GNN makes a prediction, which parts of the input graph are relevant? Ideally, the generated explanation/subgraph should lead to the same prediction outcome by the pre-trained GNN (e.g. $p\left(\tilde{y}\,|\,G^s\right)$ should be close to $p\left(\tilde{y}\,|\,G^c\right)$). In other words, a better explainer should be able to generate more compact subgraphs yet maintains the prediction accuracy while the associated explanations are fed into the pre-trained GNN. To this end, we generate the explanations for the test set based on {\em Gem}, GNNExplainer and PGExplainer, respectively. Then we use the predictions of the pre-trained GNN for the explanations to calculate the explanation accuracy.


The rationality of $K$ selection is from the size of the ground-truth structure for the synthetic datasets. We do not have the ground-truth motif for the real-world datasets, therefore we select $K$ according to the distribution of the graph size for each dataset, respectively (reported in the Appendix B). Note that a $K$ that is too small may incur meaningless explanations; for example, the aromatic group is an essential component leading to the mutagenic effect. We evaluate the explanation performance under different $K$ settings, starting from $K=15$ for Mutag and NCI1. 
\begin{table}[!t]
\caption{Explanation Accuracy on Synthetic Datasets ($\%$).}
\begin{adjustbox}{center}
\begin{small}
\setlength{\tabcolsep}{0.15em}
\begin{tabular}{ c l || c c c c c | c c c c c}
\cmidrule[1pt]{2-12}
& & \multicolumn{5}{{c}}{BA-SHAPES}  & \multicolumn{5}{{c}}{TREE-CYCLES}\\
&K&5&6&7&8&9&6&7&8&9&10\\
\cmidrule{2-12}
&{\em Gem}&$\textbf{93.4}$&$\textbf{97.1}$&$\textbf{97.1}$&$\textbf{97.1}$&$\textbf{99.3}$&$86.1$&$\textbf{87.5}$&$\textbf{92.5}$&$\textbf{93.9}$&$95.4$\\
&GNNExplainer&$82.4$&$88.2$&$91.2$&$91.2$&$94.1$&$14.3$&$46.8$&$74.6$&$91.4$&$\textbf{96.1}$\\
&PGExplainer&$71.9$&$90.7$&$92.0$&$93.3$&$94.1$&$\textbf{94.4}$&$80.6$&$77.0$&$82.4$&$89.4$\\
\cmidrule[1pt]{2-12}
\end{tabular}
\end{small} 
\end{adjustbox}
\label{tab:syn}
\vspace{-24pt}
\end{table} 

\begin{table}[!t]
\caption{Explanation Accuracy on Real-World Datasets ($\%$).}
\begin{adjustbox}{center}
\begin{small}
\setlength{\tabcolsep}{0.25em}
\begin{tabular}{ c l || c c c c | c c c c}
\cmidrule[1pt]{2-10}
&& \multicolumn{4}{{c}}{MUTAG}  & \multicolumn{4}{{c}}{NCI1}\\
&K&15&20&25&30&15&20&25&30\\
\cmidrule{2-10}
&{\em Gem-0}&$\textbf{64.0}$&$\textbf{78.1}$&$\textbf{81.0}$&$\textbf{85.0}$&$-$&$-$&$-$&$-$\\
&GNNExplainer-0&$60.0$&$67.6$&$68.9$&$75.8$&$-$&$-$&$-$&$-$\\
&PGExplainer-0&$22.5$&$38.5$&$57.6$&$72.3$&$-$&$-$&$-$&$-$\\
&{\em Gem}&$66.3$&$\textbf{78.0}$&$\textbf{82.1}$&$\textbf{83.4}$&$56.9$&$\textbf{65.3}$&$68.9$&$\textbf{72.8}$\\
&GNNExplainer&$\textbf{67.1}$&$74.9$&$75.8$&$80.9$&$\textbf{59.3}$&$61.8$&$\textbf{69.6}$&$72.0$\\
\cmidrule[1pt]{2-10}
\end{tabular}
\end{small} 
\end{adjustbox}
\vspace{-22pt}
\label{tab:real}
\end{table} 

\begin{figure}[t]
  \centering
  \includegraphics[scale = 0.02]{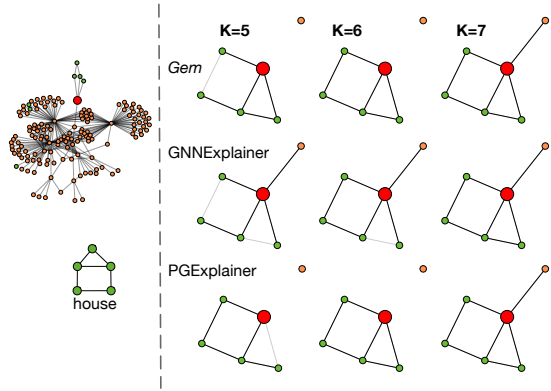}
  \caption{Explanation comparisons on BA-shapes. The ``house'' in green is the ground-truth motif that determines the node labels. The red node is the target node to be explained (better seen in color).}
  \label{fig:syn1}
  \vspace{-17pt}
\end{figure}

\subsection{Experimental Results}
\label{subsec:settings}

In what follows, we summarize the results of our experimental results and discuss our key findings. The explanation accuracy for synthetic datasets and real-world datasets in different $K$ settings are reported in Table~\ref{tab:syn} and Table~\ref{tab:real}, respectively. As shown in Table~\ref{tab:syn}, {\em Gem} consistently offers the best explanation accuracy in overall cases. In particular, {\em Gem} achieves $30\%$ improvement when $K=5$ on BA-shapes, compared with GNNExplainer. By looking at the explanations for a target node, shown in Figure~\ref{fig:syn1}, {\em Gem} can successfully identify the ``house'' motif that explains the node label (``middle-node'' in red), when $K=6$, while the GNNExplainer wrongly attributes the prediction to a node (in orange) that is out of the ``house'' motif. On Tree-cycles, GNNExplainer failed to generate effective explanations when $K<8$, while {\em Gem} and PGExplainer achieves favorable accuracy even when $K=6$. 

Note that, for the real-world datasets, there are no explicit motifs (no ground truth motifs) for classification. PGExplainer assumes $\textrm{NO}_2$ or $\textrm{NH}_2$ as the motifs for the mutagen graphs and trains an MLP for model explanation with the mutagen graphs including at least one of these two motifs. For fair comparisons, we report the results of PGExplainer following its setting reported in~\cite{luo2020parameterized} and compare them with the results of GNNExplainer and {\em Gem} when explaining on mutagen graphs, indicated as PGExplainer-0, GNNExplainer-0, and {\em Gem-0} in Table~\ref{tab:real}. As GNNExplainer and {\em Gem} can explain both classes in the dataset, we report the results of explaining the entire test set using GNNExplainer and {\em Gem} (the 4-5th rows in Table~\ref{tab:real}\footnote{The comparisons with PGExplainer on NCI1 were omitted as it fails to explain on NCI1, indicated as ``$-$'' in Table~\ref{tab:real} and Table~\ref{tab:inference}.}). For NCI1, PGExplainer fails to explain this dataset without the motif assumption, therefore, we report the results of GNNExplainer and {\em Gem}. In general, the results reported successfully verify that our proposed {\em Gem} can generate explanations that can consistently yield high explanation accuracies over all datasets.

\begin{figure}[t]
  \centering
  \includegraphics[scale = 0.19]{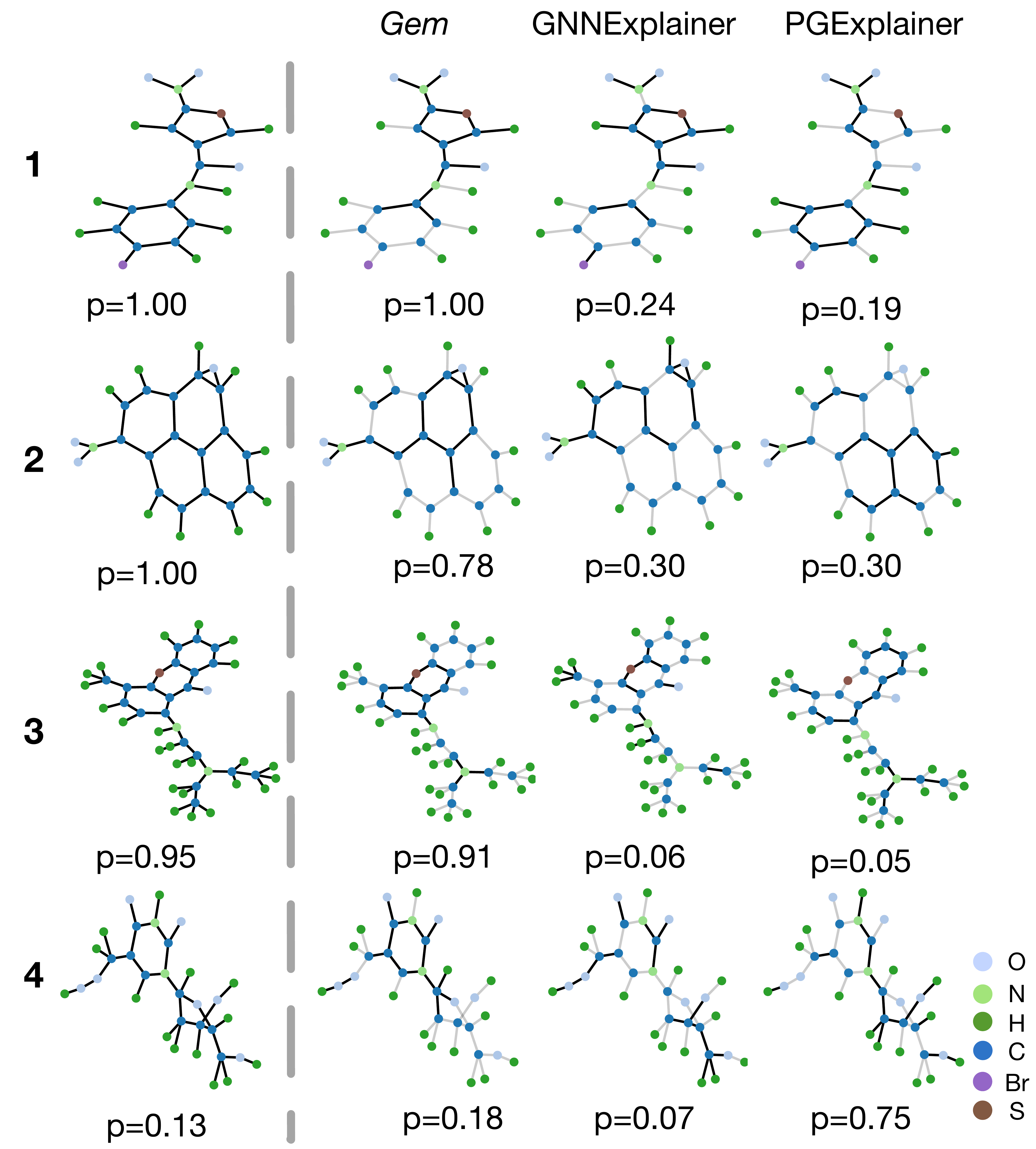}
  \vspace{-10pt}
  \caption{Explanation comparisons on Mutag. The explanation results of different methods are highlighted with black edges, where the gray edges are regarded as unimportant components for the prediction and are discarded by the explainers (better seen in color). The probability under each graph/subgraph denotes the likelihood of being classified into the ``mutagenic'' class, which is obtained by feeding the associated graph/subgraph into the pre-trained GNN.}
  \label{fig:mutag}
  \vspace{-17pt}
\end{figure}

\begin{figure}[t]
  \centering
  \includegraphics[scale = 0.28]{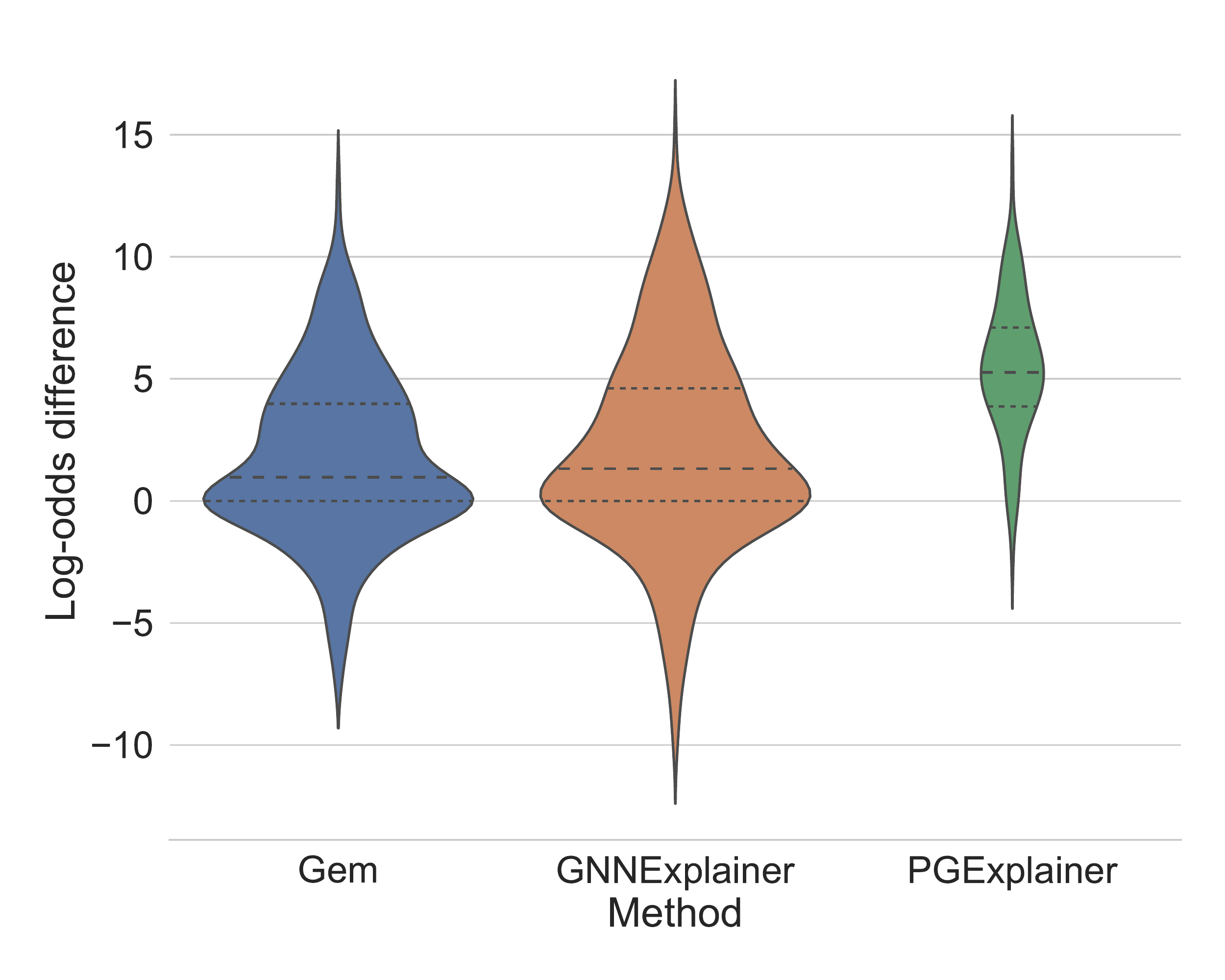}
  \vspace{-10pt}
  \caption{Log-odds difference comparisons on Mutag (more dense distribution around $0$ is better).}
  \label{fig:log_odd}
  \vspace{-15pt}
\end{figure}

To further check the interpretability of the generated explanations, we report the explanation results for Mutag in Figure~\ref{fig:mutag} ($K=15$). The first column shows the initial graphs and corresponding probabilities of being classified as ``mutagenic'' class by the pre-trained GNN, while the other columns report the explanation subgraphs. Associated probabilities belonging to the ``mutagenic'' class based on the pre-trained GNN are reported below the subgraphs. 

In the first two cases (the first two rows), {\em Gem} can identify the essential components --- the aromatic group (carbon ring) and $\textrm{NO}_2$ --- leading to their labels ( ``mutagenic''). Nevertheless, GNNExplainer either only recognizes the aromatic group (the first row) or $\textrm{NO}_2$ group (the second row), which is not sufficient to be classified into ``mutagenic'' class by the pre-trained GNN. PGExplainer focuses on identifying the pre-defined motifs --- $\textrm{NO}_2$ and $\textrm{NH}_2$. In the first two rows, we observe that PGExplainer can successfully recognize the $\textrm{NO}_2$ motifs. However, when its explanation subgraphs are fed into the pre-trained GNN, the probabilities of being classified into the correct class are quite low. In the third row of Figure~\ref{fig:mutag}, we report an instance that belongs to the ``mutagenic'' class without $\textrm{NO}_2$ or $\textrm{NH}_2$ motifs. Only {\em Gem} can recognize the essential components of classifying into ``mutagenic.'' Note that, a good explainer for a pre-trained GNN should be able to highlight the important components that lead to its predictions for any instances. In the last row of Figure~\ref{fig:mutag}, we report an instance that belongs to the ``non-mutagenic'' class. Though there are no explicit motifs for this class, {\em Gem} can successfully generate the explanation, which can be recognized as the ``non-mutagenic'' class by the target GNN, with a probability of $0.82$.

To verify the effectiveness of {\em Gem} in a more statistical view, we measure the resulting change in the pre-trained GNNs' outcome by computing the difference (the initial graph and the explanation subgraph with $K=15$ for Mutag) in log odds and investigate the distributions over the entire test set. The result on the Mutag dataset is reported in Figure~\ref{fig:log_odd} (The definition of log-odds difference and more results on other datasets are provided in Appendix B). We can observe that {\em Gem} performs consistently better than GNNExplainer and PGExplainer. Specifically, the log-odds difference of {\em Gem} is more concentrated around $0$, which indicates {\em Gem} can well capture the most relevant subgraphs towards the predictions by the pre-trained GNNs. 

\begin{table}[!t]
\caption{Inference Time per Instance (ms).}
\begin{center}
\begin{small}\small\addtolength{\tabcolsep}{-8pt}
\begin{sc}
\setlength{\tabcolsep}{0.07em}
\begin{tabular}{ c| c |c |c| c}
\toprule
Datasets & BA-shapes & Tree-cycles & Mutag & NCI1\\ 
\midrule
GNNExplainer&$265.2$ & $204.5$ & $257.6$ & $259.8$\\
PGExplainer&$6.7$ & $6.5$ & $5.5$ & $-$\\
Gem&$0.5$ & $0.5$ & $0.05$ & $0.02$\\
\bottomrule
\end{tabular}
\end{sc}
\end{small} 
\end{center}
\vspace{-25pt}
\label{tab:inference}
\end{table}

{\bf Computational performance.} PGExplainer and {\em Gem} can explain unseen instances in the inductive setting. We measure the average inference time for these two methods. As GNNExplainer explains an instance at a time, we measure its average time cost per explanation for comparisons. As reported in Table~\ref{tab:inference}, we can conclude that {\em Gem} consistently explain faster than the baselines overall. Further experiments on the efficiency evaluation, such as training time comparisons, are provided in Appendix B.



%% file: related.tex
\section{Other Related Work}

\label{sec:related}

Explanations seek the answers to the questions of ``what if'' and ``why,'' which are arguably and inherently causal. The theory of causal inference is one method by which such questions might be answered~\cite{pearl2009causality}. Recently, causal interpretability has gained increasing attention in explaining machine learning models~\cite{schwab2019cxplain,narendra2018explaining,datta2016algorithmic}. There are several viable formalisms of causality, such as Granger causality~\cite{granger1969investigating}, causal Bayesian networks~\cite{pearl1985bayesian}, and structural causal models~\cite{pearl2009causality}.

Prior works of this research line usually are designed to explain the importance of each component of a neural network on its prediction. Chattopadhyay {\em et al.}~\cite{chattopadhyay2019neural} proposed an attribution method based on the first principles of causality, particularly the Structural Causal Model and $\mathbf{do}(\cdot)$ calculus. Schwab~{\em et al.}~\cite{schwab2019cxplain} framed the explanation task for deep learning models on images as a causal learning task, and proposed a causal explanation model that can learn to estimate the feature importance towards its prediction. Schwab~{\em et al.}'s proposal is built upon the notion of Granger causality. While such methods can provide meaningful explanations for deep models on images, they cannot be directly applied to interpret graph neural networks, due to the inherent property of graph representations. 


%% file: conclusion.tex
\section{Conclusion}
\label{sec:conclusion}

In sum, we devised a new framework {\em Gem} for explaining graph neural networks that use the first principles of Granger causality.  {\em Gem} has several advantages over existing work: it is model-agnostic, compatible with any graph neural network models without any prior assumptions on the graph learning tasks, can generate compact subgraphs, causing the outputs of the pre-trained GNNs very quickly after training. We show that causal interpretability could contribute to explaining and understanding graph neural networks. We believe this could be a fruitful avenue of future research that helps better understand and design graph neural networks. 


%% file: ack.tex
\section{Acknowledgement}
This project is supported by the Internal Research Fund at The Hong Kong Polytechnic University P0035763.

%% file: appendix.tex
\section{A Unified Framework for Explaining Graph Neural Networks}
\label{sec:unified}

We adopt the unified framework for explaining neural networks defined in~\cite{lundberg2017unified}. {\em Additive feature attribution methods} have an explanation model that is a linear function of binary variables. Existing methods unified in this framework provide meaningful explanations for deep models on images. To specifically interpret graph neural networks on graph-structured data, we adapt the definition of the additive property for graph data attribution as follows:

\begin{equation}\label{eq:add_attr}
\epsilon(A^s)=\psi_0 +\sum\psi_{ij}\times A^s_{ij}
\end{equation}

where $A^s$ is the adjacency matrix of the explanation, $A^s_{ij}\in\{0,1\}$ is a binary variable representing the existence of an edge. 

In this section, we analyze that GNNExplainer~\cite{ying2019gnnexplainer}, PGExplainer~\cite{luo2020parameterized} and our explainer, {\em Gem}, all fall into the framework of additive feature attribution methods. In what follows, we analyze, these methods for explaining GNNs essentially solve the same optimization problem with different approximation methods. 

In general, to identify a compact subgraph that is important for the GNN's prediction, the explanation problem can be formulated as following optimization problem based on mutual information:
\begin{equation}\label{eq:gnnex}
\begin{aligned}
\max_{G^s\subseteq G^c}  \quad &\mathbf{MI}\left(Y, G^s\right)\\
=\max_{G^s\subseteq G^c} \quad & \left(H(Y)-H(Y|G=G^s)\right)\\
s.t. \quad & |G^s|<K
\end{aligned}
\end{equation}
where $K$ is a constraint on $G^s$'s size for a compact explanation.

When treating $G^s\subseteq G^c$ as a graph random variable and with convexity assumption, Eq.~(\ref{eq:gnnex}) can be approximated as solving following optimization problem:

\begin{equation}\label{eq:gnnex_rel}
\begin{aligned}
\min_{G^s\subseteq G^c} H(Y|G=\mathbb{E}(G^s)).
\end{aligned}
\end{equation}

To make the optimization tractable, GNNExplainer relaxes the integer constraint on the adjacency matrix and searches for an optimal fractional adjacency matrix. In particular, GNNExplainer factorizes the explanation into a multivariate Bernoulli distribution, formulated as $P(G_s)=\Pi_{(i,j)\in G^s}A^s_{ij}$. For explanations on graph structures, an explanation is selected based on the values of entries in the solution $A^s$. GNNExplainer uses mean-field optimization to solve $\min_{A^s}H(Y|G=\mathbb{E}(G^s))$. Therefore, the weights $\psi_{ij}$ of GNNExplainer under the unified framework are chosen based on the solution $A^s$.

{\em Gem} approximates the solution of the optimization problem from a causal perspective and utilize a graph generative model to generate explanations in an inductive setting. Specifically, {\em Gem} assumes that the selections of edges from the $G^c$ are conditionally independent to each other. With the assumption of conditional independence, we estimate the edge importance based on the notion of Granger causality. These estimations then are used as the ground-truth to train our explanation model. In particular, we use an auto-encoder architecture utilizing graph convolutions~\cite{kipf2016variational,li2018multi,li2018learning}, due to its simplicity and expressiveness. Following the instantiation of graph auto-encode~\cite{kipf2016variational},  we use the reparameterization trick for training, our objective is approximated as $\min_{Z\sim\mathcal{N}(\mu,\sigma)}H(Y|G=G^s)$. The weights $\psi_{ij}$ of {\em Gem} under the unified framework are then chosen based on the output of the graph generative model.

PGExplainer factorizes the explanation graph as $P(G_s)=\Pi_{(i,j)\in G^s}P(A^s_{ij})$, where $P(A^s_{ij})$ is instantiated with the Bernoulli distribution $A^s_{ij}\sim \mathbf{Bern}(\theta_{ij})$. 
PGExplainer utilizes the reparameterization trick and approximate the objective as $\min_{\epsilon\sim\mathbf{Uniform}(0,1)}H(Y|G=G^s)$. To collectively explain multiple instances, PGExplainer adopts a multi-layer perception (MLP) to learn to generate explanations from the target GNN, taking the node embeddings generated from the target GNN as the input. The weights $\psi_{ij}$ of PGExplainer under the unified framework are chosen based on the output of the MLP. 

\section{Further Implementation Details and More Experimental Results}
\label{implementations}

{\bf Datasets.} BA-shapes was created with a base Barabasi-Albert (BA) graph containing $300$ nodes and $80$ five-node ``house''-structured network motifs. Tree-cycles was built with a base $8$-level balanced binary tree and $80$ six-node cycle motifs. The statistics of four datasets are presented in Table~\ref{tab:data_sta}. Note that, we report the average number of nodes and the average number of edges over all the graphs for the real-world datasets. Table~\ref{tab:data_acc} reports the model accuracy on four datasets, which indicates that the models to be explained are performed reasonably well. 

The motif statistics of Mutag are reported in Table~\ref{tab:mutag_motif}. We found that there are $32\%$ of non-mutagenic graphs containing at least $\textrm{NO}_2$ or $\textrm{NH}_2$ motifs, which indicates that the assumption made by PGExplainer might not be reasonable. 

\begin{table}[!ht]
\caption{Data Statistics of Four Datasets.}
\begin{center}
\begin{small}\small\addtolength{\tabcolsep}{-3pt}
\begin{sc}
\setlength{\tabcolsep}{0.03em}
\begin{tabular}{ c| c |c |c| c}
\toprule
Datasets & BA-shapes & Tree-cycles & Mutag & NCI1\\ 
\midrule
\#graphs&$1$ & $1$ & $4,337$ & $4,110$\\
\#nodes&$700$&$871$&$29$&$30$\\
\#edges&$4,110$&$1,950$&$30$&$32$\\
\#labels&$4$&$2$&$2$&$2$\\
\bottomrule
\end{tabular}
\end{sc}
\end{small}	
\end{center}
\label{tab:data_sta}
\end{table}

\begin{table}[!ht]
\caption{Model Accuracy of Four Datasets ($\%$).}
\begin{center}
\begin{small}\small\addtolength{\tabcolsep}{-3pt}
\begin{sc}
\setlength{\tabcolsep}{0.03em}
\begin{tabular}{ c| c |c |c| c}
\toprule
Datasets & BA-shapes & Tree-cycles & Mutag & NCI1\\ 
\midrule
Accuracy&$94.1$ & $97.1$ & $88.5$ & $78.6$\\
\bottomrule
\end{tabular}
\end{sc}
\end{small}	
\end{center}
\label{tab:data_acc}
\end{table}

\begin{table}[!ht]
\caption{Motif Statistics on Mutag (Number of Instances Containing Associated Motif).}
\begin{center}
\begin{small}\small\addtolength{\tabcolsep}{-2pt}
\begin{sc}
\setlength{\tabcolsep}{0.01em}
\begin{tabular}{ c| c |c}
\toprule
Motif & $\textrm{NO}_2$& $\textrm{NH}_2$\\ 
\midrule
Mutagenic &$596$ & $501$  \\
Non-Mutagenic &$97$ & $251$\\
\bottomrule
\end{tabular}
\end{sc}
\end{small}	
\end{center}
\label{tab:mutag_motif}
\end{table}

Unless otherwise stated, all models, including GNN classification models and our explainer, are implemented using PyTorch~\footnote{https://pytorch.org} and trained with Adam optimizer. All the experiments were performed on a NVIDIA GTX 1080 Ti GPU with an Intel Core i7-8700K processor. 

\begin{figure}[ht!]
  \centering
  \includegraphics[scale = 0.06]{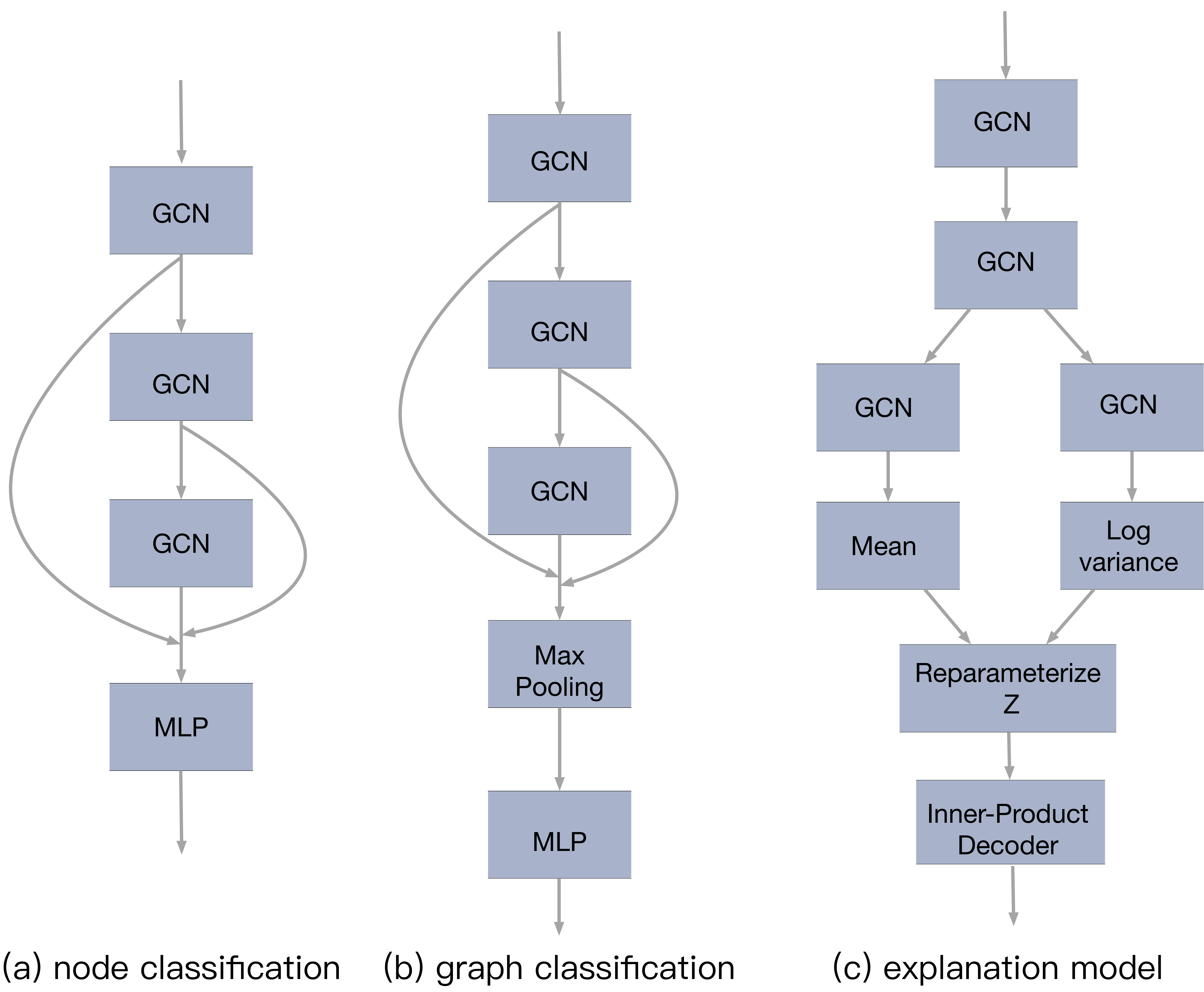}
  \caption{Model architectures.}
  \label{fig:model-arch}
\end{figure}

For GNN classification models, we use the same setting as GNNExplainer. Specifically, for node classification, we apply three layers of GCNs with output dimensions equal to $20$ and perform concatenation to the output of three layers, followed by a linear transformation to obtain the node label. For graph classification, we employ three layers of GCNs with dimensions of $20$ and perform global max-pooling to obtain the graph representations. Then a linear transformation layer is applied to obtain the graph label. Figure~\ref{fig:model-arch} (a) and ~\ref{fig:model-arch} (b) are the model architectures for node classification and graph classification, receptively. 

\begin{figure}[!ht]%
 \centering
 \subfloat[Mutag.]{\includegraphics[scale = 0.25]{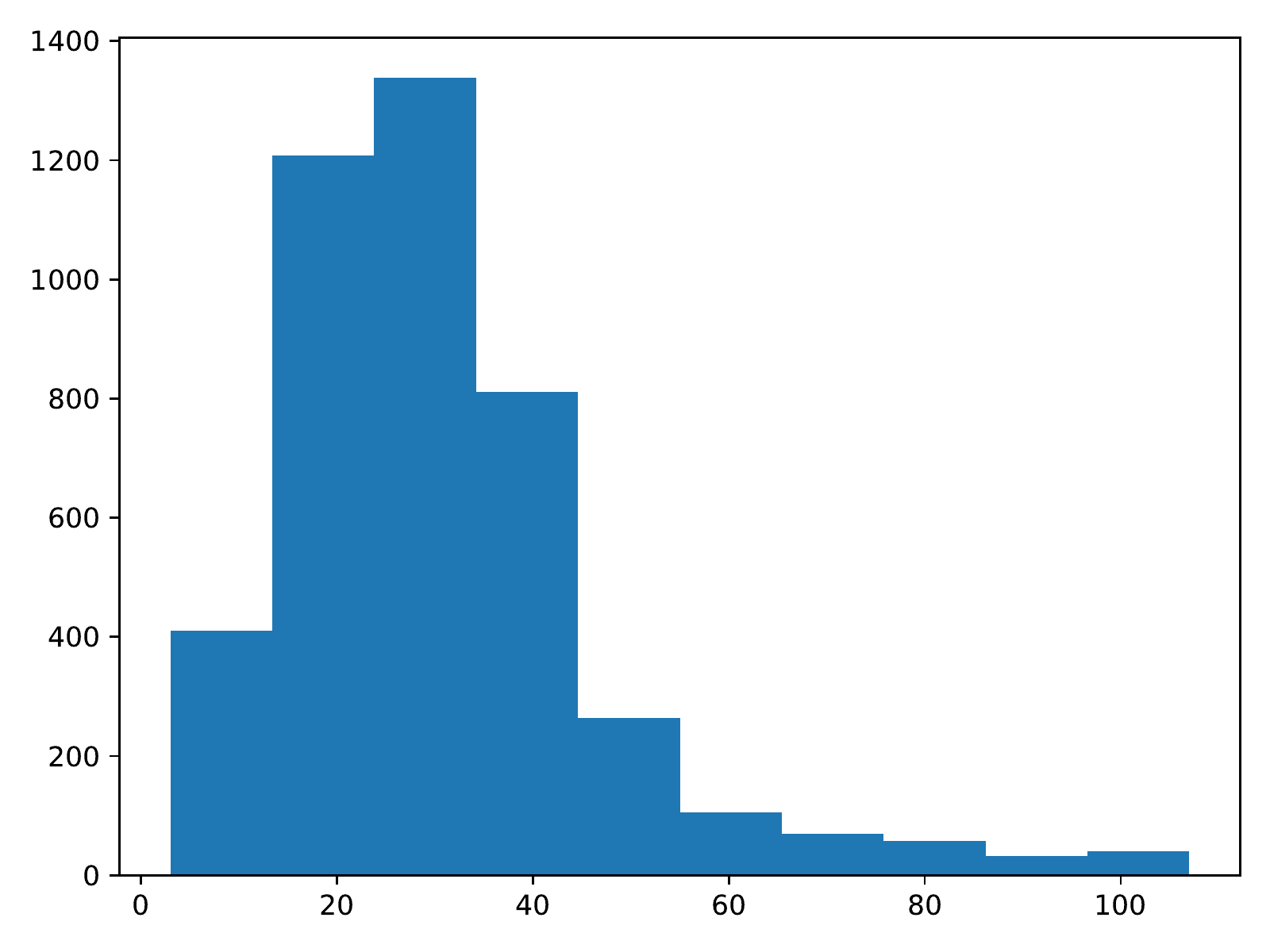}\label{fig:mutag_hist}}
 \subfloat[NCI1.]{\includegraphics[scale = 0.25]{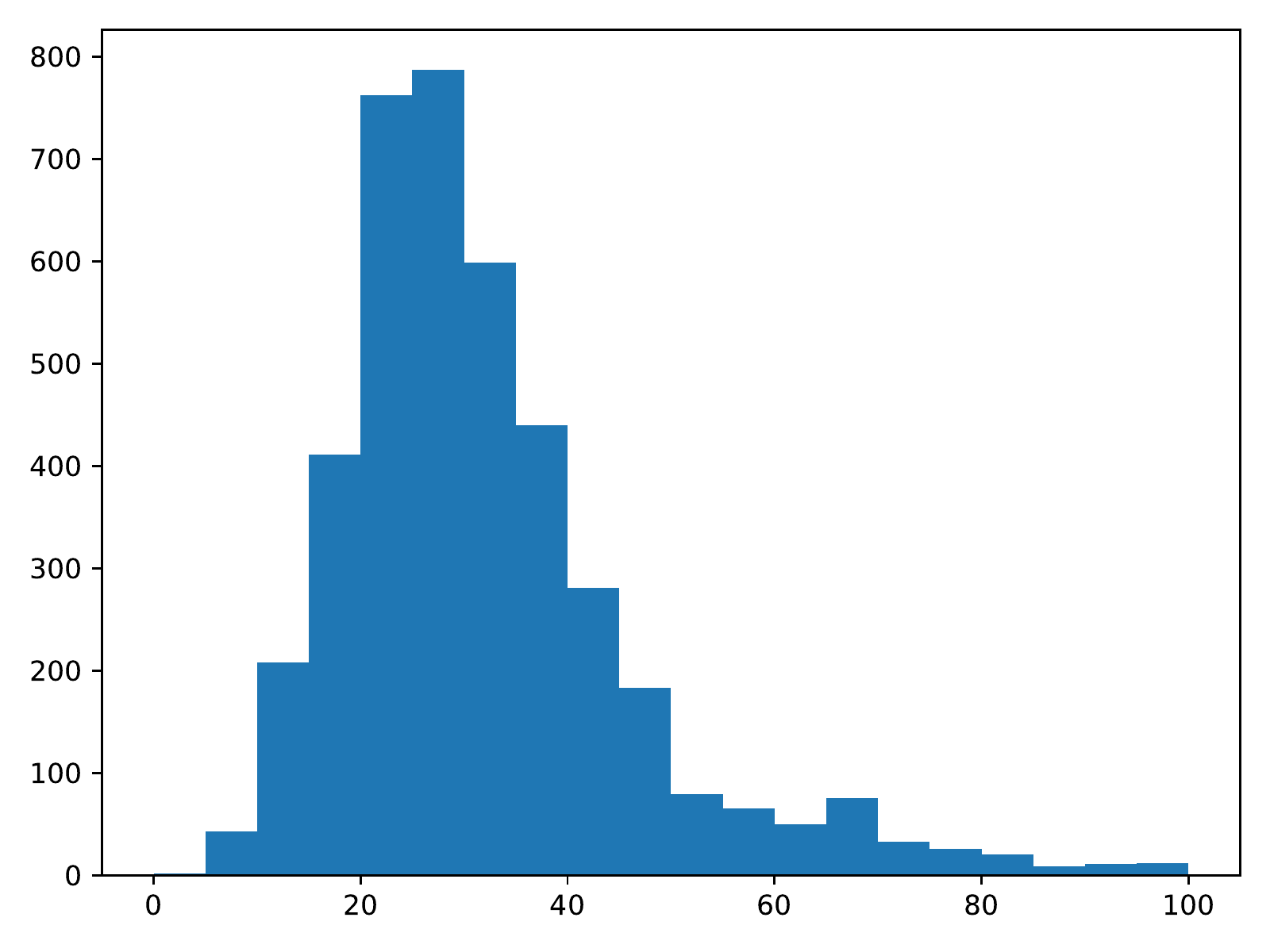}\label{fig:NCI1_hist}}
 \vspace{9pt}
 \caption{The distribution of the graph size on real-world datasets.}%
 \label{fig:scal}%
\end{figure}

Figure~\ref{fig:model-arch} (c) depicts the model architecture of {\em Gem} for generating explanations. We use mean square error as the loss for training {\em Gem}. In particular, it was optimized using Adam optimizer with a learning rate of $0.01$ and $0.001$ for explaining graph and node classification model, respectively. We train at batch size $32$ for $100$ epochs. Table~\ref{tab:stats} shows the detailed data splitting for model training, testing, and validation. Note that both classification models and our explanation models use the same data splitting. Figure~\ref{fig:scal} depicts the distribution of the graph size on two real-world datasets in terms of the number of edges.

\begin{table}[!t]
\caption{Data Splitting for Four Datasets.}
\begin{center}
\begin{small}\small\addtolength{\tabcolsep}{-5pt}
\begin{sc}
\setlength{\tabcolsep}{0.07em}
\begin{tabular}{ c| c |c |c}
\toprule
Datasets & \#of Training & \#of Testing & \#of Validation\\ 
\midrule
Ba-shapes&$300$ & $50$ & $50$\\
Tree-cycles&$270$ & $45$ & $45$\\
Mutag&$3,468$ & $434$ & $434$\\
NCI1&$3,031$ & $410$ & $411$\\
\bottomrule
\end{tabular}
\end{sc}
\end{small}	
\end{center}
\label{tab:stats}
\end{table}

{\bf Log-odds difference.} We measure the resulting change in the pre-trained GNNs' outcome by computing the difference in log odds and investigate the distributions over the entire test set. The log-odds difference is formulated as:
\begin{equation}\label{eq:log}
\Delta\textrm{log-odds}=\textrm{log-odds}\left(p(G^c)\right)-\textrm{log-odds}\left(p(G^s)\right)
\end{equation}
where $\textrm{log-odds}(p)=\mathbf{log}\left(\frac{p}{1-p}\right)$, and $p(G^c)$ and $p(G^s)$ are the outputs of the pre-trained GNN. 

\begin{figure}[!ht]%
 \centering
 \subfloat[BA-shapes.]{\includegraphics[scale = 0.28]{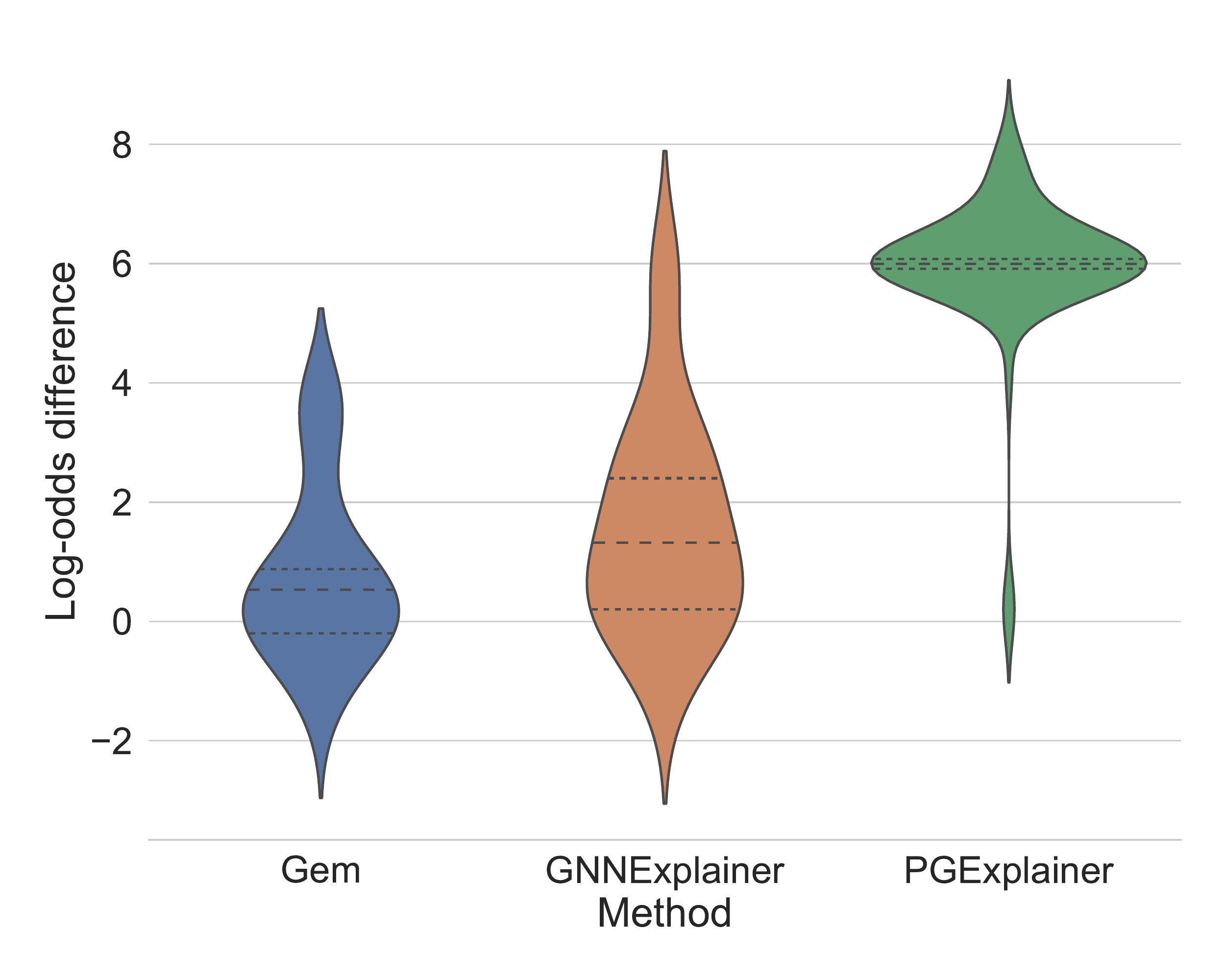}\label{fig:ba-shapes_log_odd}}\\
 \subfloat[Tree-cycle.]{\includegraphics[scale = 0.28]{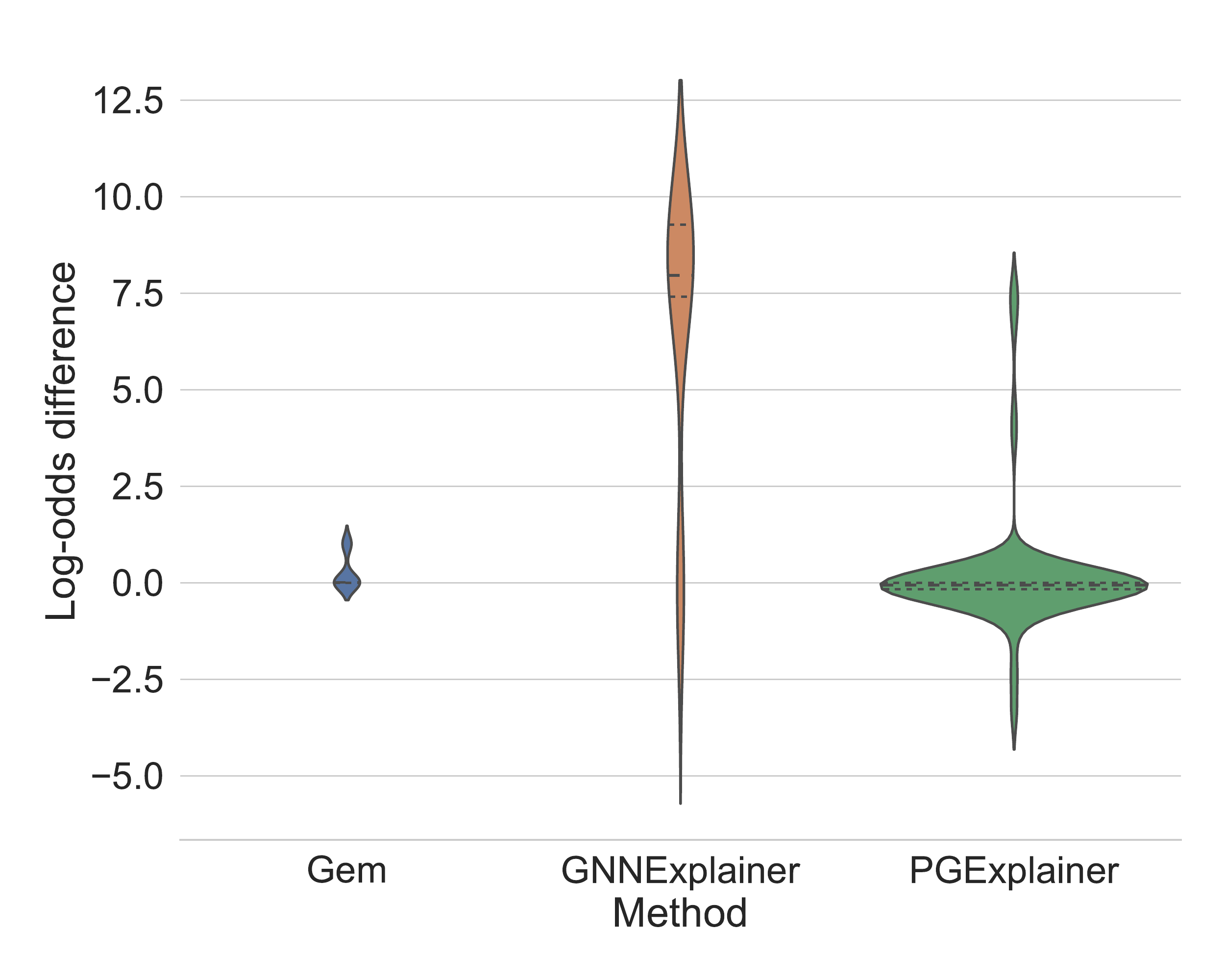}\label{fig:tree-cycle_log_odd}}\\
 \subfloat[NCI1.]{\includegraphics[scale = 0.28]{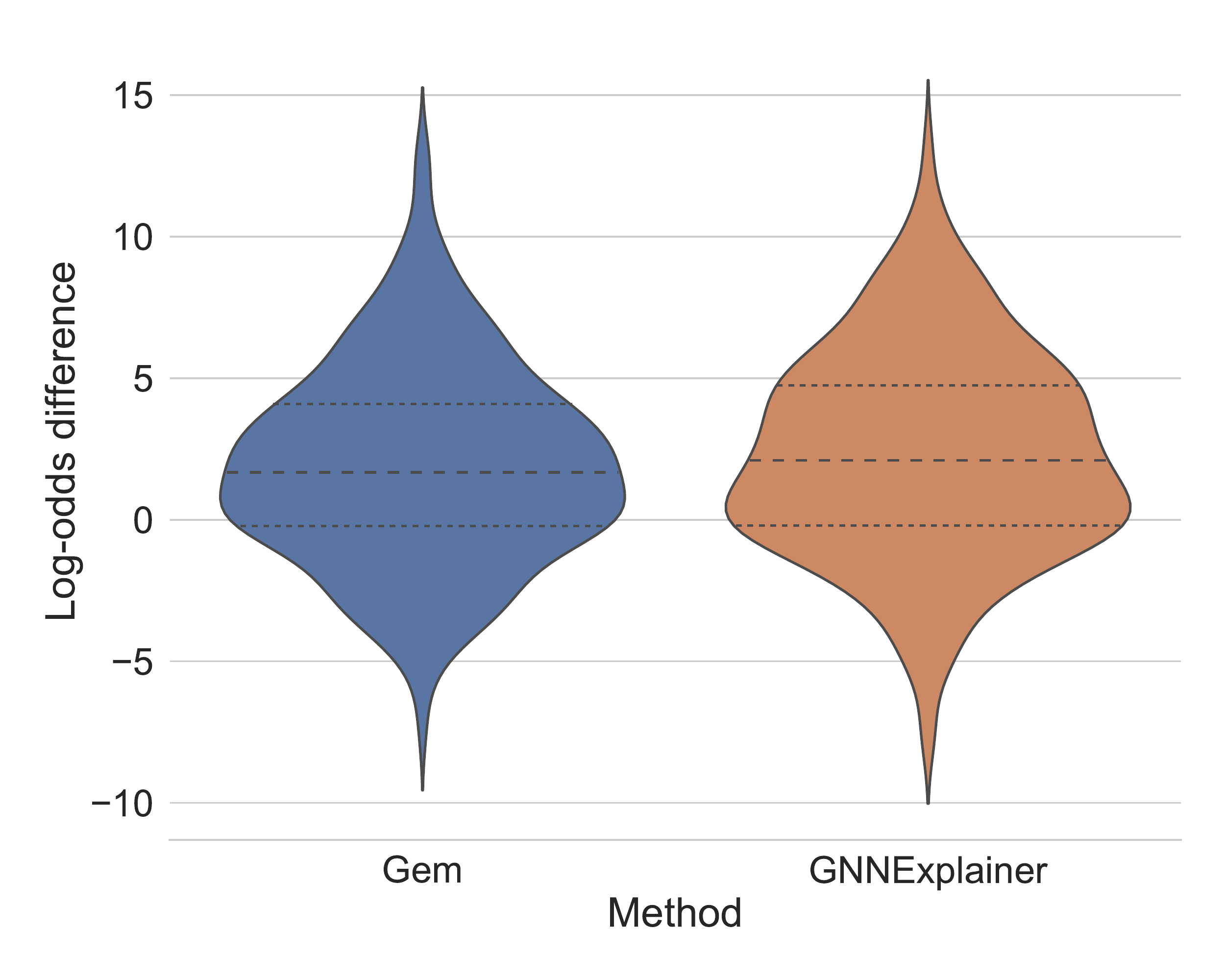}\label{fig:nci1_log_odd}}
 \caption{Log-odds difference comparisons on other datasets (more dense distribution around $0$ is better).}%
 \label{fig:log_odd_app}%
\end{figure}

The results on the other datasets are reported in Figure~\ref{fig:log_odd_app}. We can observe that {\em Gem} performs consistently better than GNNExplainer and PGExplainer in general. Specifically, the log-odds difference of {\em Gem} is more concentrated around $0$, especially on the BA-shapes dataset, which indicates {\em Gem} has a better computation accuracy on identifying the most relevant subgraphs towards the predictions by the pre-trained GNNs. 

\begin{figure}[!ht]%
 \centering
 \subfloat[BA-shapes.]{\includegraphics[scale = 0.18]{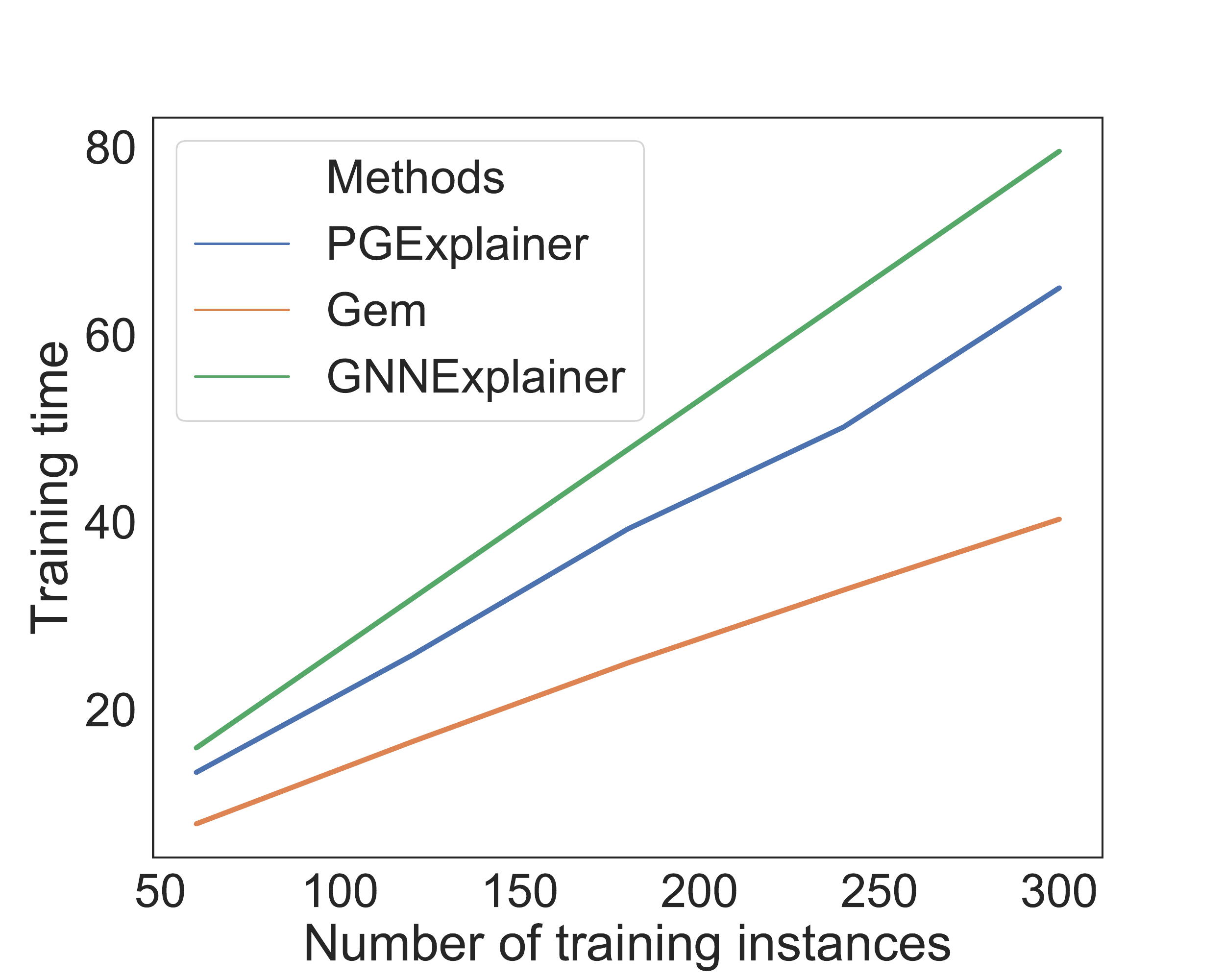}\label{fig:syn1_train_time}}
 \subfloat[Tree-cycles.]{\includegraphics[scale = 0.18]{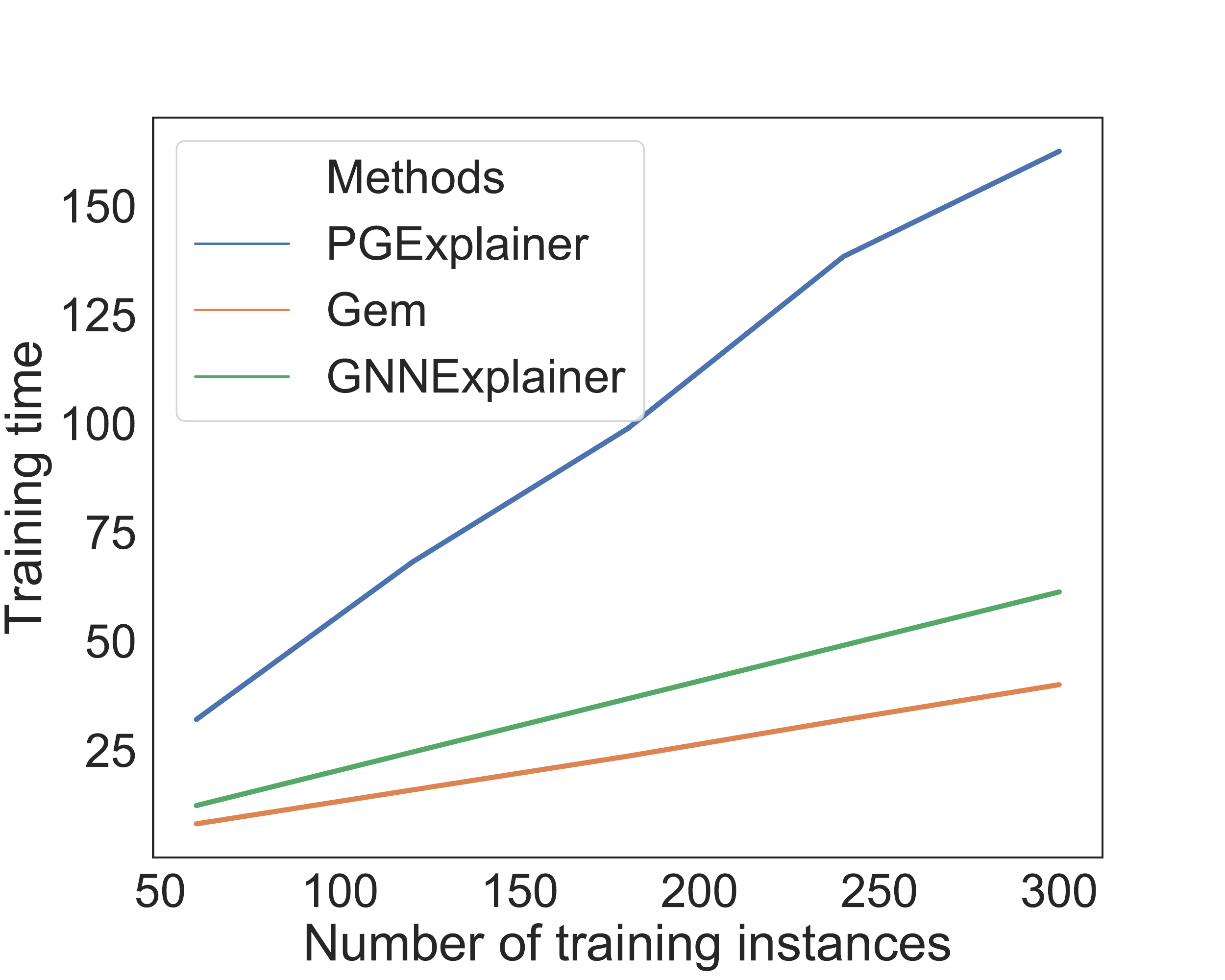}\label{fig:syn4_train_time}}\\
 \subfloat[Mutag.]{\includegraphics[scale = 0.18]{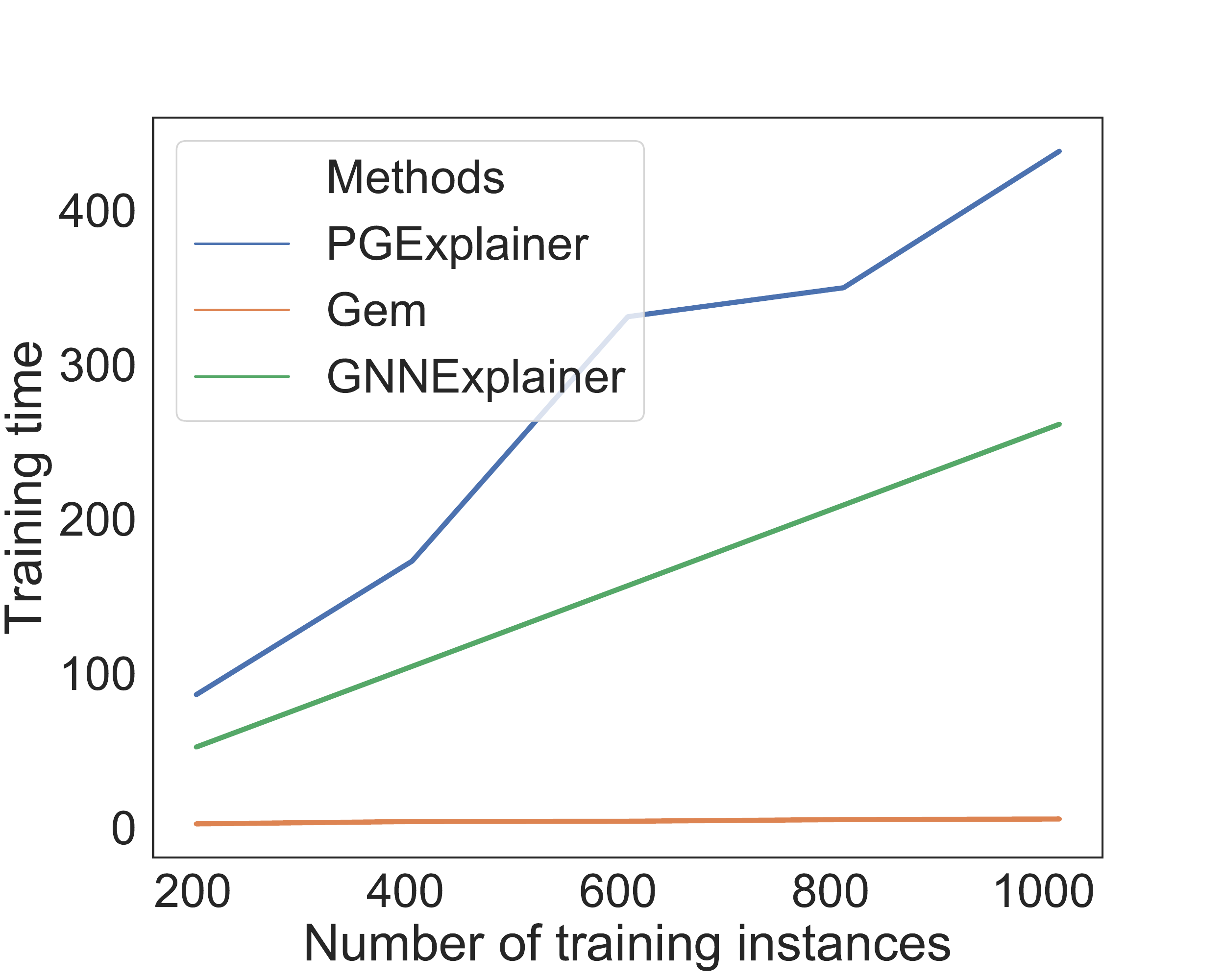}\label{fig:mutag_train_time}}
 \subfloat[NCI1.]{\includegraphics[scale = 0.18]{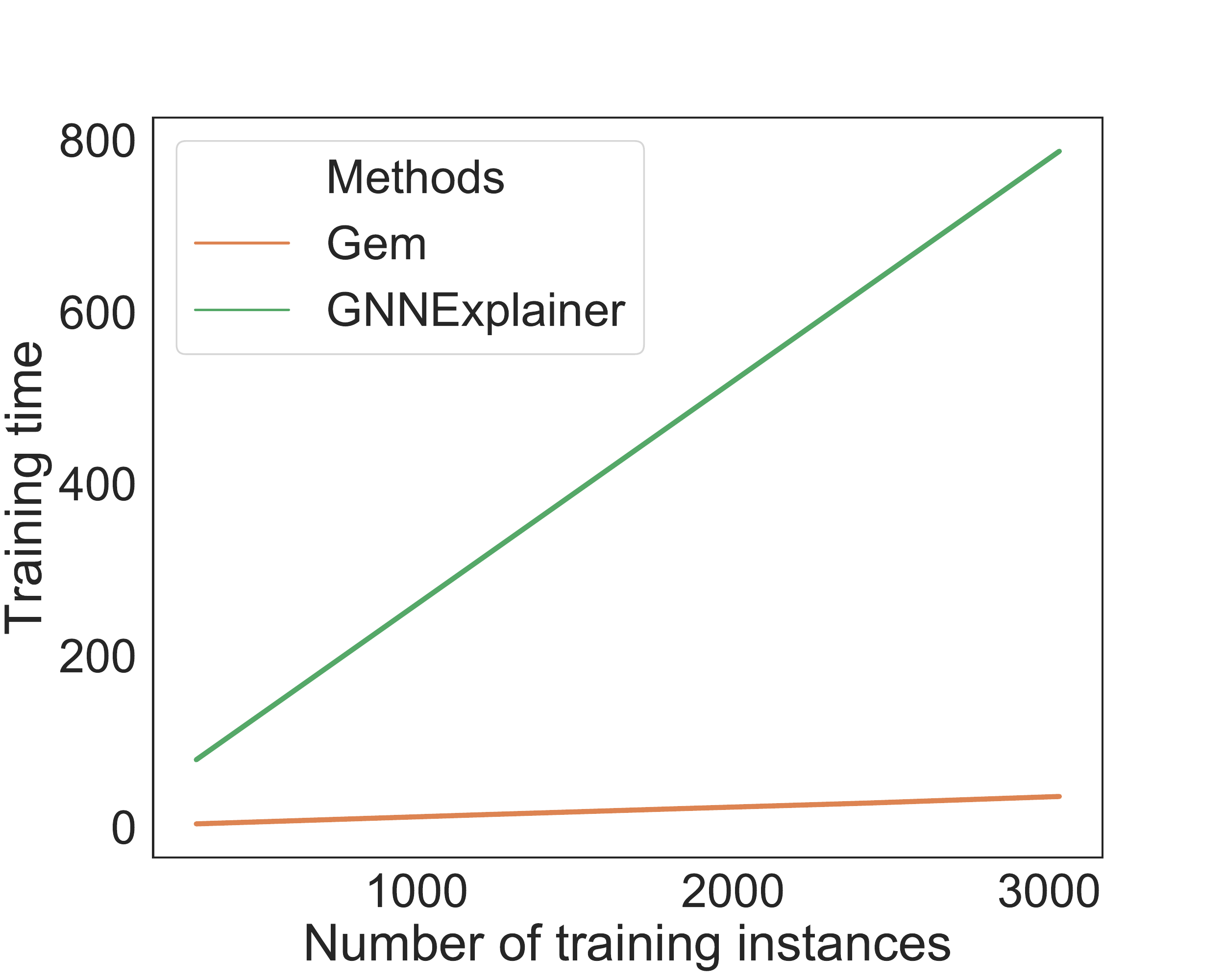}\label{fig:nci1_train_time}}
 \caption{Training time comparisons.}%
 \label{fig:training_time}%
\end{figure}

{\bf Computational performance.} For {\em Gem}, we measure the time for distilling the ``ground-truth'' explanation and the time for training the explainer with a varying number of training instances. As for GNNExplaner, we measure the overall time cost of explanations with a varying number of target instances. The training time of PGExplainer and the explanation time of GNNExplainer are measured by reusing the source code released by the authors. Figure~\ref{fig:training_time} report the computational performance for four datasets. Note that, as Figure~\ref{fig:training_time} shows, it takes $5.67$ seconds to train Mutag with $1015$ samples, which is quite fast (small but not zero). The distillation process is specifically designed for our framework and only occurs before our explainer training, which can be thought of as a ``pre-processing'' step. In other words, the time is a one-time cost. The computation time cost of the distillation process is provided in Table~\ref{tab:distill}. We can conclude that {\em Gem} amortizes the explanation cost by training a graph generator to generate explanations for any given instances and consistently explain faster than the baseline overall.

\section{Distillation Process}
\label{alg:distill}

We are given a pre-trained classification model, denoted as $f\left(\cdot\right)$, and the training set that is used to train $f\left(\cdot\right)$. {\sc\bf Procedure Distillation Process} presents the detailed algorithm of generating the ground-truth explanation $G^s$ for a training instance.

\begin{algorithm}[!ht]
\caption{{\sc\bf Distillation Process}: Distill the top-k most relevant edges for each computation graph}
\label{alg:disll_alg}
\begin{algorithmic}[1]
\REQUIRE{Given the computation graph $G^c = (V^c, A^c, X^c)$.}
\ENSURE{Calculate the model error of the computation graph $\delta_{G^c}$ according to Eq.~(4)}
\FOR{edge $e_j\in\,G^c$}
	\STATE{Calculate the causal contribution $\Delta{\scriptscriptstyle{\delta,\,e_j}}$ according to Eq.~(1) -- Eq.~(5).}	
\ENDFOR
\STATE{Remove edges with the least casual contribution and re-calculate the causal contribution of the generated subgraph.}
\STATE{$E^c_\text{sorted}\leftarrow$ sort the edges in ascending order based on the causal contributions.}
\STATE{Initialize $G^s \leftarrow G^c$}
\FOR{$\forall$ edge $e_j\in\mathit{E}^c_\text{sorted}$}
	\STATE{Calculate the model error of the subgraph, denoted as $\delta_{G^s\setminus\left\{e_j\right\}}$ according to Eq.~(3) and Eq.~(5).}
	\STATE{$G^{s0} \leftarrow G^s\setminus\left\{e_j\right\}$}
	\IF{$G^{s0}$ must be connected}
		\STATE{$G^{s0} \leftarrow $ largest component of $G^{s0}$}
	\ENDIF
	\IF{$\delta_{G^{s0}} > \delta_{G^s}$}
		\STATE{$e_j$.weight $\leftarrow \delta_{G^s} - \delta_{G^{s0}}$}
	\ELSE
		\STATE{$G^s \leftarrow G^{s0}$}
	\ENDIF
\ENDFOR
\STATE{Distill the subgraph with top-$k$ most relevant edges.}
\STATE{$\mathit{E}^s \leftarrow $ edges of $G^s$}
\STATE{$\mathit{E}^s_\text{sorted} \leftarrow $ sort $\mathit{E}^s$ in ascending order by weight}
\FOR{$\forall$ edge $e_j \in \mathit{E}^s_\text{sorted}$}
	\IF{\# of edge in $G^s \geq k $}
		\STATE{$G^s \leftarrow G^s\setminus\left\{e_j\right\}$}
		\IF{$G^s$ must be connected}
			\STATE{$G^s \leftarrow $ largest component of $G^s$}
		\ENDIF
	\ENDIF
\ENDFOR
\STATE{return $G^s$}

\end{algorithmic}
\end{algorithm}

The quality of the distilled ground truth provides a basis for our explanation performance. We also provide the performance evaluation of the distillation to see how well the distillation process works. We put the accuracy and associate time cost in Table~\ref{tab:distill}.

\begin{table}[!t]
\caption{Distillation Performance on Real-World Datasets.}
\begin{adjustbox}{center}
\begin{small}
\setlength{\tabcolsep}{0.25em}
\begin{tabular}{ c l || c c c c | c c c c}
\cmidrule[1pt]{2-10}
&& \multicolumn{4}{{c}}{MUTAG}  & \multicolumn{4}{{c}}{NCI1}\\
\cmidrule{2-10}
&K&15&20&25&30&15&20&25&30\\
&Accuracy(\%)&$90.7$&$94.2$&$95.2$&$95.8$&$82.0$&$91.7$&$96.8$&$97.1$\\
\cmidrule{2-10}
&\# of instance &200&400&600&1000&300&600&1200&3000\\
&Time(s)&$18$&$36$&$54$&$90$&$18$&$37$&$74$&$186$\\
\cmidrule[1pt]{2-10}
\end{tabular}
\end{small} 
\end{adjustbox}
\label{tab:distill}
\end{table}